\pdfoutput=1

\documentclass[a4paper,fleqn]{cas-dc}

\usepackage[numbers]{natbib}

\usepackage{amsmath,amsfonts,bm}

\def\eqref#1{equation~\ref{#1}}

\def\1{\bm{1}}

\def\vtheta{{\bm{\theta}}}
\def\va{{\bm{a}}}

\def\vc{{\bm{c}}}
\def\vd{{\bm{d}}}
\def\ve{{\bm{e}}}
\def\vf{{\bm{f}}}

\def\vs{{\bm{s}}}

\def\vu{{\bm{u}}}
\def\vv{{\bm{v}}}

\def\vx{{\bm{x}}}

\def\vz{{\bm{z}}}

\def\mA{{\bm{A}}}

\def\mF{{\bm{F}}}

\def\mL{{\bm{L}}}

\def\mP{{\bm{P}}}

\def\mW{{\bm{W}}}
\def\mX{{\bm{X}}}
\def\mY{{\bm{Y}}}
\def\mZ{{\bm{Z}}}

\def\mSigma{{\bm{\Sigma}}}

\DeclareMathAlphabet{\mathsfit}{\encodingdefault}{\sfdefault}{m}{sl}
\SetMathAlphabet{\mathsfit}{bold}{\encodingdefault}{\sfdefault}{bx}{n}

\def\gH{{\mathcal{H}}}

\def\gL{{\mathcal{L}}}

\newcommand{\R}{\mathbb{R}}

\newcommand{\sigmoid}{\sigma}

\usepackage{url}
\usepackage{adjustbox}
\usepackage{dsfont}

\ExplSyntaxOn
\cs_set:Npn \__make_tbl_caption:nn #1#2
{
  \l_tbl_align_tl
  \skip_vertical:N \l_tbl_abovecap_skip
  {\parbox{\linewidth}
    {\rightskip=0pt\sffamily\small\textbf{\color{scolor}#1}\par#2\par\vskip4pt }}
  \skip_vertical:N \l_tbl_belowcap_skip
}
\ExplSyntaxOff

\makeatletter
\AtEndDocument{\clearpage\immediate\write\@auxout{\string\csxdef{lastpage}{\the\numexpr\value{page}-1\relax}}}
\makeatother
\begin{document}
\let\WriteBookmarks\relax
\def\floatpagepagefraction{1}
\def\textpagefraction{.001}

\shorttitle{Arbitrary-Scale Anisotropic 3D MRI Super-Resolution with Gaussian Prototypes}

\shortauthors{}

\title [mode = title]{Arbitrary-Scale Anisotropic 3D MRI Super-Resolution with Gaussian Prototypes}

\author[1,2]{Aditya Rastogi}
\fnmark[1]
\ead{aditya@ccibonn.ai}
\ead[url]{https://ccibonn.ai/authors/aditya/}
\credit{Formal analysis, Conceptualization, Investigation, Methodology, Writing -- original draft}

\author[1,2]{Ramon Leiser}
\fnmark[1]
\credit{Data curation, Validation, Visualization, Writing -- original draft}

\author[1,2]{Prateek Singh}
\credit{Data curation, Validation, Writing -- original draft}

\author[1,2]{Martha Foltyn-Dumitru}
\credit{Formal analysis, Supervision, Writing -- review and editing}

\author[3]{Martin Bendszus}
\credit{Data curation, Writing -- review and editing}

\author[4]{Fabian Bamberg}
\credit{Data curation for NAKO, Writing -- review and editing}

\author[5,6]{Klaus Maier-Hein}
\credit{Data curation, Supervision, Writing -- review and editing}

\author[7]{Kai Schlamp}
\credit{Writing -- review and editing}

\author[1,8]{Alexander Radbruch}
\credit{Writing -- review and editing}

\author[1,2,5]{Gianluca Brugnara}
\credit{Formal analysis, Supervision, Writing -- review and editing}

\author[1,2,5]{Philipp Vollmuth}
\cormark[1]

\ead{philipp.vollmuth@ukbonn.de}
\ead[URL]{https://ccibonn.ai/authors/philipp/}
\credit{Conceptualization, Investigation, Writing -- review and editing, Supervision, Resources, Project administration}

\affiliation[1]{organization={University of Bonn, University Hospital Bonn, Department of Neuroradiology},
                addressline={Venusberg-Campus 1}, 
                city={Bonn},
                postcode={53127}, 
                state={North Rhine-Westphalia},
                country={Germany}}

\affiliation[2]{organization={Division of Computational Radiology and Clinical AI},
                addressline={Baunscheidstraße 17}, 
                city={Bonn},
                postcode={53113}, 
                state={North Rhine-Westphalia},
                country={Germany}}

\affiliation[3]{organization={Department of Neuroradiology, Heidelberg University Hospital},
                addressline={Im Neuenheimer Feld 400}, 
                city={Heidelberg},
                postcode={69120}, 
                state={Baden-Württemberg},
                country={Germany}}

\affiliation[4]{organization={Department of Diagnostic and Interventional Radiology, Faculty of Medicine, University Medical Center Freiburg, University of Freiburg},
               addressline={Hugstetter Str. 55}, 
                city={Freiburg},
                postcode={79106}, 
                state={Baden-Württemberg},
                country={Germany}}

\affiliation[5]{organization={German Cancer Research Centre (DKFZ)},
               addressline={Im Neuenheimer Feld 280}, 
                city={Heidelberg},
                postcode={6912}, 
                state={Baden-Württemberg},
                country={Germany}}

\affiliation[6]{organization={Pattern Analysis and Learning Group, Department of Radiation Oncology, Heidelberg University Hospital},
               addressline={Im Neuenheimer Feld 410}, 
                city={Heidelberg},
                postcode={6912}, 
                state={Baden-Württemberg},
                country={Germany}}

\affiliation[7]{organization={Department of Diagnostic and Interventional Radiology, Thoraxklinik at Heidelberg University Hospital},
                addressline={Röntgenstraße 1}, 
                city={Heidelberg},
                postcode={69126}, 
                state={Baden-Württemberg},
                country={Germany}}

\affiliation[8]{organization={German Center for Neurodegenerative Diseases (DZNE)},
               addressline={Venusberg-Campus 1}, 
                city={Bonn},
                postcode={53127}, 
                state={North Rhine-Westphalia},
                country={Germany}}

\fntext[1]{Equal contribution}


\begin{abstract}
Reducing MRI acquisition time can improve patient comfort, reduce motion artefacts, and increase scanner throughput. Clinical protocols often achieve shorter acquisitions by sacrificing spatial resolution, particularly along the slice axis. Super-resolution can reconstruct higher-resolution images, but existing methods are commonly specialised to a particular scale or acquisition geometry, limiting their use across heterogeneous clinical protocols. We present GSMR3D, a continuous volumetric model that reconstructs arbitrary isotropic and anisotropic resolutions using a single checkpoint per MRI contrast. GSMR3D combines a learned dictionary of anisotropic 3D Gaussian prototypes with scale-conditioned frequency encoding and a measurement-consistency projection, keeping reconstructions consistent with the acquired low-resolution data. We train separate models on 3,731 contrast-enhanced T1-weighted and 30,254 T1-weighted brain MRI volumes from clinical and population-based cohorts, respectively. We evaluate GSMR3D against geometry-specific specialists across isotropic, in-plane, and through-plane super-resolution using reconstruction quality and downstream imaging tasks, including brain tumour segmentation, brain-metastasis detection, and brain parcellation. GSMR3D outperforms specialists at isotropic $2\times$ and achieves better LPIPS, DISTS, and FID across all geometries, although specialists retain higher PSNR in anisotropic settings. GSMR3D more faithfully preserves glioma segmentations across all geometries, achieves the highest lesion sensitivity on an independent brain-metastasis cohort, and achieves the highest brain-parcellation agreement for isotropic and in-plane reconstruction. Reconstruction takes approximately 6\,s per volume, adding little overhead relative to potential acquisition-time savings. GSMR3D provides a flexible approach to MRI scan-time reduction using one model per contrast across acquisition geometries while preserving information relevant to downstream imaging tasks.

\end{abstract}


\begin{graphicalabstract}
\includegraphics[width=\textwidth]{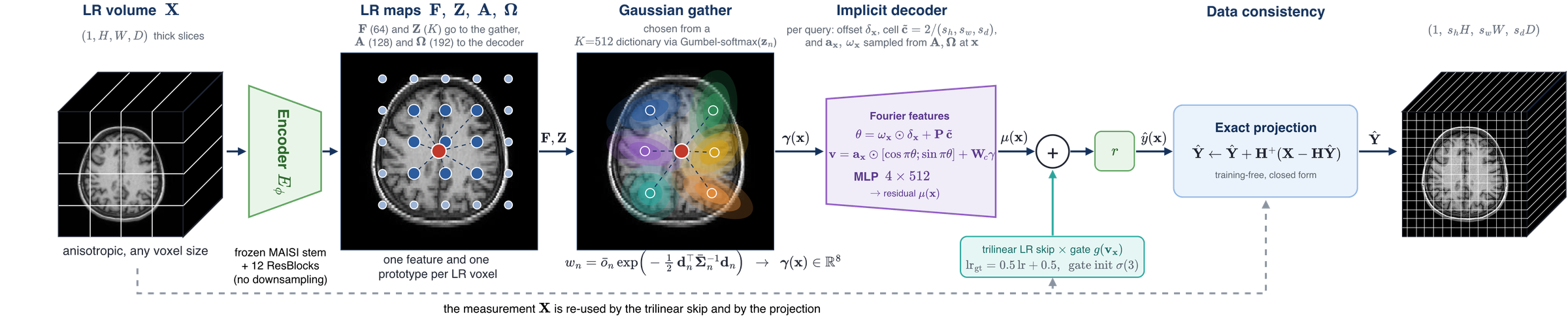}
\end{graphicalabstract}

\begin{highlights}

    \item One model handles arbitrary-scale, isotropic and anisotropic MRI super-resolution.

    \item Learned 3D Gaussian prototypes enable continuous volumetric reconstruction.

    \item A fused CUDA implementation reconstructs a whole MRI volume in about 6s.

    \item GSMR3D best preserves glioma segmentation and brain-metastasis lesion sensitivity.

\end{highlights}

\begin{keywords}
Magnetic resonance imaging \sep super-resolution \sep arbitrary-scale reconstruction \sep anisotropic imaging \sep Gaussian representations \sep medical image analysis
\end{keywords}

\maketitle


\section{Introduction}
\label{sec:intro}

Higher-resolution (HR) MRI can reveal finer anatomical detail, but acquiring smaller voxels generally requires longer scan times, increasing susceptibility to motion artefacts and reducing patient tolerance \citep{khateriMRISuperResolutionDeep2025,andre2015motion}. Clinical protocols therefore often trade spatial resolution for acquisition speed, particularly in the through-plane direction, producing volumes with high in-plane resolution but thick slices. Such anisotropic scans are common in clinical practice, but can obscure small structures and pathology, while segmentation and morphometry tools used to derive quantitative biomarkers often perform less reliably on them \citep{billot2023synthseg,morell-ortegaCAHALClinicallyApplicable2026,iglesias2023synthsr}. Limited spatial resolution is also a challenge for low- and ultra-low-field MRI, where improved accessibility and lower cost are accompanied by coarser resolution \citep{lauPushingLimitsLowcost2023,deleeuwdenbouterDeepLearningbasedSingle2022}. Super-resolution could therefore enable faster MRI acquisitions while retaining clinically relevant spatial detail.

Super-resolution (SR) offers a computational solution for recovering HR detail from rapidly acquired low-resolution (LR) scans \citep{zhangMRImageSuperResolution2021,kimLightweightBrainMR2024}. Most MRI SR methods, however, are designed for specific upsampling factors and acquisition geometries. Isotropic 3D SR, in-plane SR, and reconstruction of thick through-plane slices are therefore commonly treated as separate problems with specialised models for each \citep{khateriMRISuperResolutionDeep2025,peng2020saint,zhao2021smore}. This is restrictive for clinical MRI, where voxel spacing varies between scanners, sequences, and hospitals, and a model trained for one geometry may not directly apply when the factor or direction of subsampling changes. Arbitrary-scale SR addresses this limitation by allowing a single model to reconstruct images at different target resolutions. One approach is to model the image as a continuous representation, as introduced for natural images by LIIF and LTE \citep{chen2021liif,lee2022lte} and later extended to 3D MRI by ArSSR \citep{wu2023arssr}. Another approach for arbitrary-scale SR is through generative models. Diffusion models have been adapted to arbitrary-scale MRI SR, either with implicit representations \citep{hanArbitraryScaleSuperresolution2024} or through reconstruction processes that accommodate different target resolutions \citep{wang_DiffusionbasedArbitraryscale_2026}. Unconditional generative priors can likewise be combined with different degradation operators at inference \citep{song2022solving}. However, independent scaling along all three spatial axes has received considerably less attention. A practical MRI SR model should be able to handle different scale factors and acquisition geometries without requiring a separate model for each setting, while remaining competitive with geometry-specific models.

A separate challenge concerns the fidelity of the reconstructed anatomy. Generative models, particularly diffusion-based approaches, can produce visually convincing high-frequency detail. However, these details need not correspond to the anatomy that was actually acquired. Recent studies have shown that such reconstruction errors can alter anatomical structures and affect measurements derived from SR images \citep{morell-ortegaCAHALClinicallyApplicable2026,kim_MPFlowMultimodal_2026}. This is particularly important in medical imaging, where reconstructed volumes may subsequently be used for segmentation, morphometry, or other quantitative analyses. Diffusion-based reconstruction also requires iterative sampling, which can make volumetric inference computationally expensive. However, MRI offers an important advantage in this setting because an approximation of the acquisition process can be incorporated into the reconstruction through measurement consistency. This measurement consistency can be used to ensure that the reconstructed HR image remains compatible with the acquired LR data under the assumed degradation model \citep{song2022solving,hu2024restoration}.

To address these challenges, we introduce GSMR3D, a continuous 3D MRI SR model that reconstructs arbitrary isotropic and anisotropic resolutions from a single checkpoint. Inspired by Gaussian splatting \citep{kerbl2023gs} and its adaptation to arbitrary-scale image SR \citep{hu2025gaussiansr,chen2025generalized}, GSMR3D uses anisotropic Gaussians to learn a continuous representation of the scan from discrete LR voxels. It learns a shared dictionary of 3D Gaussian prototypes whose anisotropic covariances allow their support to vary across spatial directions. Rather than scattering each Gaussian over an increasing number of output voxels as the target resolution grows, GSMR3D gathers contributions from a fixed LR neighbourhood for each HR query. This keeps the computational cost of the gather per query independent of the upsampling factor and makes it practical for volumetric SR. The gathered features are combined with a scale-conditioned frequency encoding, and a measurement projection makes the reconstruction consistent with the acquired LR image.

We evaluate GSMR3D on T1-weighted and contrast-enhanced T1-weighted brain MRI across isotropic, in-plane, and through-plane acquisition geometries. We compare it with geometry-specific specialists, a diffusion prior with data consistency, and a residual CNN trained across the same scale distribution. We additionally evaluate whether the reconstructed images preserve information relevant to downstream analysis using glioma segmentation, healthy-brain parcellation, and lesion-level analysis in an independent brain-metastasis cohort. A single GSMR3D checkpoint per contrast outperforms the isotropic specialist across the evaluated image-quality metrics and achieves lower LPIPS, DISTS, and FID than geometry-specific specialists across all evaluated geometries. Importantly, GSMR3D more faithfully preserves glioma segmentations across all geometries and achieves the highest brain-metastasis lesion sensitivity, including settings in which specialist models achieve higher PSNR. With a custom CUDA implementation of the Gaussian gather, GSMR3D reconstructs a volume in approximately 6\,s, around 13 times faster than the evaluated diffusion model. Together, these results demonstrate that GSMR3D can reconstruct MRI across diverse acquisition geometries while preserving clinically relevant information, supporting its potential for faster MRI acquisitions that could reduce scan times and improve patient throughput.

\section{Related work}
\label{sec:related}

\paragraph{MRI SR.}
Deep learning has substantially advanced MRI SR, with convolutional and attention-based models learning to recover HR images from synthetically degraded inputs \citep{lyuMRISuperResolutionEnsemble2020,zhangMRImageSuperResolution2021,kimLightweightBrainMR2024}. Several methods further exploit the structure of specific acquisition geometries. SAINT performs 2D SR on sagittal and coronal slices and fuses the reconstructed views \citep{peng2020saint}. SMORE and ECLARE use self-supervision, learning from high-resolution in-plane information within the input volume for through-plane reconstruction \citep{zhao2021smore,remedios2026eclare}. SynthSR maps heterogeneous clinical scans to a 1\,mm isotropic T1 volume \citep{iglesias2023synthsr}, while CAHAL uses experts conditioned on image resolution and anisotropy \citep{morell-ortegaCAHALClinicallyApplicable2026}. These methods demonstrate the benefits of accounting for acquisition geometry, while motivating models that can accommodate a broader range of resolutions and anisotropies within a single framework.

\paragraph{Arbitrary-scale SR.}

Several approaches have been developed for arbitrary-scale SR. For natural images, LIIF represents images as continuous functions queried at arbitrary coordinates \citep{chen2021liif}, LTE augments this representation with local frequency information \citep{lee2022lte}, and CiaoSR aggregates neighbouring features through implicit attention \citep{cao2023ciaosr}. In MRI, ArSSR extends continuous implicit SR to arbitrary isotropic 3D factors \citep{wu2023arssr}, while CuNeRF enables zero-shot arbitrary-scale reconstruction from individual volumes \citep{chen2023cunerf}. ASSRDM combines continuous representations with diffusion \citep{hanArbitraryScaleSuperresolution2024}, whereas PRDDiff uses progressive $k$-space diffusion for arbitrary in-plane SR \citep{wang_DiffusionbasedArbitraryscale_2026}. GSMR3D instead conditions on the target resolution independently along each axis and combines scale-conditioned frequency encoding with learned anisotropic Gaussian support.

\paragraph{Gaussian representations for SR.}
Gaussian representations provide another approach to arbitrary-scale SR. Gaussian splatting represents scenes using anisotropic Gaussian primitives rasterised onto the image plane \citep{kerbl2023gs}. GaussianSR and GSASR adapt Gaussian representations to isometric arbitrary-scale 2D SR by predicting image-conditioned Gaussians and rendering them onto the target grid \citep{hu2025gaussiansr,chen2025generalized}. ContinuousSR observed that predicted covariances concentrate around a limited set of shapes and uses this structure as a Deep Gaussian Prior \citep{peng2026pixel}, motivating our shared prototype dictionary. Extending scatter-based rendering to volumetric SR is computationally expensive because each Gaussian can cover an increasing number of output voxels as the target resolution grows. Inspired by \citet{hu2025gaussiansr}, GSMR3D instead gathers from a fixed 27-voxel LR neighbourhood for each HR query, retaining continuous anisotropic Gaussian weighting with a fixed number of candidate contributions per query.

\paragraph{Gaussian and generative priors in medical imaging.}
Gaussian representations have been explored for MRI reconstruction from undersampled $k$-space \citep{peng2025three}, fetal slice-to-volume reconstruction \citep{danneckerFastExplicitSlicetoVolume}, and 2D MRI SR \citep{phymrisr2026}. PhyMRI-SR is particularly related, deriving a fixed covariance dictionary from Gaussians fitted to HR images and subsequently using it for 2D SR. In contrast, GSMR3D jointly learns a 3D dictionary of anisotropic Gaussian prototypes and gathers rather than scatters their contributions. Diffusion-based priors provide an alternative approach to reconstruction with data consistency \citep{song2022solving,chungMRImageDenoising2023,lee2023improving}, but require iterative inference and can introduce anatomically plausible but incorrect detail \citep{morell-ortegaCAHALClinicallyApplicable2026,kim_MPFlowMultimodal_2026}. GSMR3D instead produces a deterministic reconstruction followed by a single projection under the assumed degradation operator. Appendix~\ref{app:related} provides further discussion.


\providecommand{\vpi}{{\bm{\pi}}}
\providecommand{\vomega}{{\bm{\omega}}}
\providecommand{\vphi}{{\bm{\varphi}}}
\providecommand{\vgamma}{{\bm{\gamma}}}
\providecommand{\vdelta}{{\bm{\delta}}}
\providecommand{\vell}{{\bm{\ell}}}
\providecommand{\mOmega}{{\bm{\Omega}}}
\providecommand{\softplusop}{\operatorname{softplus}}
\providecommand{\GS}{\operatorname{GumbelSoftmax}}
\providecommand{\MLP}{\operatorname{MLP}}
\newcommand{\lrf}{\operatorname{lr}}
\providecommand{\cell}{\vc}

\section{Method}
\label{sec:method}

\begin{figure*}[t]
  \centering
  \includegraphics[width=\linewidth]{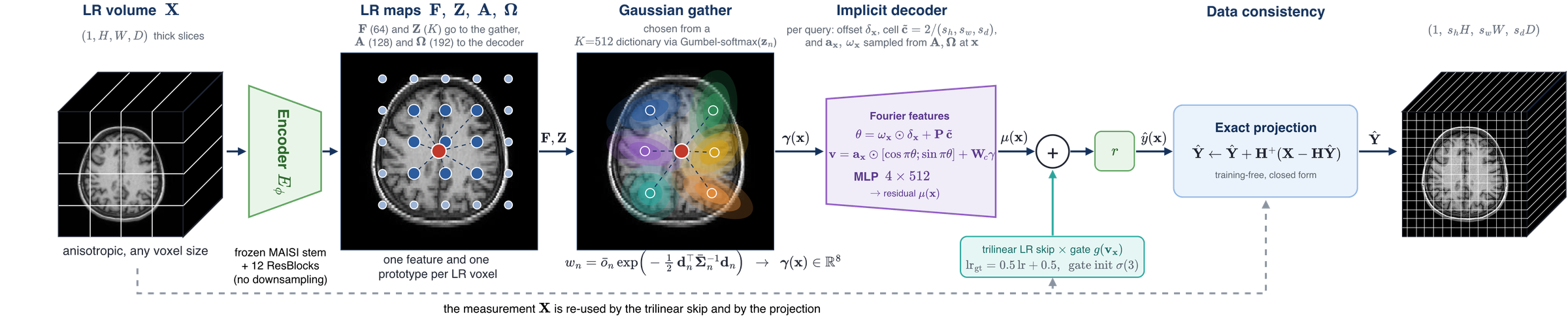}
 \caption{\textbf{GSMR3D overview.} The LR volume $\mX$ is encoded into features $\mF$, prototype logits $\mZ$, and coefficient and frequency maps $\mA$ and $\mOmega$ respectively. For each HR query $\vx$, features from a fixed $3{\times}3{\times}3$ LR neighbourhood are gathered using anisotropic Gaussians selected from a learned $K{=}512$ prototype dictionary. The gathered features are combined with scale-conditioned Fourier features and decoded into a residual, which is combined with a gated trilinear LR skip and refined. A final training-free projection enforces measurement consistency under the assumed degradation operator, producing the SR reconstruction $\hat{\mY}$ at the desired resolution.}
  \label{fig:architecture}
\end{figure*}

To perform arbitrary and anisotropic SR, GSMR3D represents the input LR scan as a continuous volumetric representation. As Fig.~\ref{fig:architecture} summarises, inference consists of two main steps: (a) feature extraction and Gaussian-prototype selection, and (b) query-based rendering. First, the encoder extracts LR features used to obtain per-voxel logits over a shared dictionary of anisotropic 3D Gaussian prototypes, together with the coefficient and frequency maps of the local Fourier decoder. This step is independent of the desired output resolution. Second, query points are defined on the target grid, and information is gathered from the LR neighbourhood using the selected Gaussian prototypes. The gathered features are combined with a scale-conditioned frequency encoding and decoded into an intensity residual, which is combined with a gated skip connection. Finally, a projection enforces consistency with the LR measurement under the assumed degradation operator.

\subsection{Problem formulation}
\label{sec:formulation}

Let $\mY\in\R^{H'\times W'\times D'}$ be an HR volume and $\mX\in\R^{H\times W\times D}$ its LR observation generated by an anisotropic degradation $\gH_{\vs}$ with nominal per-axis scale factors $\vs=(s_h,s_w,s_d)$:
\begin{equation}
  \mX = \gH_{\vs}(\mY)
  \label{eq:degradation}
\end{equation}
We seek a continuous reconstruction function
\begin{equation}
  \hat \mY(\vx) = f_\theta\big(\mX,\;\vx,\;\cell\big)
  \label{eq:implicit}
\end{equation}
which maps a query coordinate $\vx\in[-1,1]^3$ and its target voxel size (``cell'') $\cell$ to an intensity.

We normalise each spatial axis to $[-1,1]$. For an axis with $N$ voxels, the voxel centres are $x_i=-1+(2i{+}1)/N$, $i=0,\dots,N{-}1$. The target voxel size is represented by $\cell=(2/H',\,2/W',\,2/D')$, providing the decoder with the target resolution along each axis.

\subsection{Architecture overview}
\label{sec:overview}

GSMR3D consists of four main components. (i)~The \textbf{encoder} $E_\phi$ extracts an LR feature map $\mF$, from which three convolutional heads predict prototype logits $\mZ$, and coefficient and frequency maps $\mA$ and $\mOmega$ respectively, all on the LR grid. (ii)~The \textbf{Gaussian prototype dictionary} contains $K$ learnable anisotropic 3D Gaussians. For each HR query coordinate $\vx$, the logits determine the Gaussian associated with each voxel in its $3{\times}3{\times}3$ LR neighbourhood. The resulting Gaussian weights $\tilde w_n$ gather the neighbouring features into a local context vector $\vgamma(\vx)$. (iii)~The \textbf{implicit decoder} extends LTE's frequency representation to 3D. Locally sampled coefficients and frequencies are combined with the relative query position and target voxel size $\cell$ to construct a scale-conditioned Fourier representation. This is fused with $\vgamma(\vx)$ to obtain the per-query feature $\vv_{\vx}$, which an MLP maps to an intensity correction $\mu(\vx)$. (iv)~The \textbf{residual connection} combines $\mu(\vx)$ with the trilinearly interpolated LR intensity $\lrf_{\mathrm{gt}}(\vx)$ through a learned gate $g(\vx)$, followed by a refinement head $r$. The individual components are described below, with the flowchart shown in Fig.~\ref{fig:details}.

\subsection{Encoder}
\label{sec:encoder}
The encoder uses the first encoder block of the MAISI 3D VAE \citep{guo2024maisi}, as released with MAISI-v2 \citep{zhao2026maisi}. This block consists of a $1\!\to\!64$-channel convolutional layer with padding to preserve the spatial dimensions and is kept frozen during training. Its output is passed through a trainable adapter comprising $N_a{=}12$ residual blocks (\texttt{ResBlock3D}, 64 channels), producing the feature map $\mF$. Both the frozen block and the adapter operate at the original LR resolution, so $\mF$ remains on the $H{\times}W{\times}D$ grid.

Three $3^3$ convolutional heads are applied to $\mF$. A $64\!\to\!K$ head produces the prototype logits $\mZ$ used for Gaussian prototype selection, while two convolutional heads produce the coefficient map $\mA\in\R^{128\times H\times W\times D}$ and frequency map $\mOmega\in\R^{192\times H\times W\times D}$ used by the decoder. Since these maps depend only on the LR input, they can be computed once and reused for different target resolutions. The adapter and all three heads are trained end-to-end, while the MAISI block remains frozen. 

\subsection{Gaussian prototype dictionary and gather-based aggregation}
\label{sec:dictionary}

\paragraph{Dictionary parameterisation.}
The Gaussian dictionary contains $K{=}512$ learnable anisotropic 3D prototypes shared across all volumes. Each prototype $k$ is parameterised by a lower-triangular Cholesky factor $\mL_k$ and opacity $o_k$:
\begin{equation}
\begin{gathered}
\mL_k=\begin{pmatrix}\ell^{11}_k&0&0\\ \ell^{21}_k&\ell^{22}_k&0\\ \ell^{31}_k&\ell^{32}_k&\ell^{33}_k\end{pmatrix}, \\
\mSigma_k=\mL_k\mL_k^{\!\top}\succ0
\end{gathered}
\label{eq:cholesky}
\end{equation}
$\sigmoid$ is the sigmoid function. The diagonal entries of $\mL_k$ are parameterised as $\softplusop(\cdot)+10^{-4}$, ensuring that $\mSigma_k$ remains positive definite. The opacities are learned directly and initialised at $o_k{=}\sigmoid(1)\approx0.73$. The blended opacity $\bar o_j$ in Eq.~\eqref{eq:blend} is clamped to $[0,1]$.

The prototypes are initialised as axis-aligned Gaussians, with per-axis standard deviations taking eight values between $0.2$ and $3.0$, giving $8^3{=}512$ isotropic and anisotropic initial shapes. The scales are expressed in normalised coordinates, where one LR voxel spans $(2/H,2/W,2/D)$. All entries of $\mL_k$ are subsequently learned, allowing the Gaussian shapes and orientations to adapt while remaining positive definite by construction.

\paragraph{Per-voxel prototype selection.}
A prototype head $h_Z$ produces logits $\vz_j\in\R^K$ for each LR voxel $j$, which are converted to selection weights $\vpi_j$ using Gumbel-softmax with temperature $\tau = 1$. The selection is relaxed during training. At inference, the Gumbel noise is omitted and $\vpi_j=\operatorname{softmax}(\vz_j)$, so each LR voxel receives a fixed, deterministic convex blend of prototypes. The Gaussian associated with voxel $j$ is obtained by combining prototypes in Cholesky space, with the same weights applied to their opacities:

\begin{equation}
\begin{gathered}
\bar\ell^{\,aa}_j
=\softplusop\left(\sum_k \pi_{jk}\ell^{aa}_k\right)+10^{-4},\\
\bar\ell^{\,ab}_j
=\sum_k \pi_{jk}\ell^{ab}_k,\quad a>b, \\
\quad
\bar o_j
=\operatorname{clamp}\left(\sum_k \pi_{jk}o_k,\,0,\,1\right)
\end{gathered}
\label{eq:blend}
\end{equation}
Applying softplus to the blended diagonal entries ensures that $\bar{\mL}_j$ is a valid Cholesky factor and $\bar{\mSigma}_j=\bar{\mL}_j\bar{\mL}_j^{\!\top}$ remains positive definite. 

\paragraph{Gather-based aggregation.}
Conventional Gaussian splatting scatters each Gaussian to output locations within its support \citep{kerbl2023gs,chen2025generalized}. In volumetric SR, a fixed spatial support can therefore cover an increasing number of output voxels as the target resolution increases. GSMR3D instead gathers from a fixed $3{\times}3{\times}3$ LR neighbourhood for each HR query, keeping the number of candidate contributions per query fixed at 27 regardless of the target scale. Ablation studies support the choice of $K{=}512$ prototypes and a $3{\times}3{\times}3$ gather neighbourhood as a balance between reconstruction quality and computational cost (Appendix~\ref{sec:ablations}).

For a query coordinate $\vx$, let $v(\vx)$ denote its containing LR voxel with $\vx^\ast$ its centre location. The LR anchor coordinates in its $3{\times}3{\times}3$ neighbourhood are:
\begin{equation}
\begin{gathered}
\vx_n=\operatorname{clamp}\left(\vx^\ast+\Delta_n,[-1,1]\right),\qquad \\
\Delta_n=(i,j,k)\odot\left(\tfrac{2}{H},\tfrac{2}{W},\tfrac{2}{D}\right),
\qquad i,j,k\in\{-1,0,1\}
\end{gathered}
\end{equation}
From each neighbour $n$, we take the first $C_g{=}8$ channels of the encoder feature map, $\vf_n$, as a compact feature representation for the Gaussian gather, and obtain its Gaussian parameters $(\bar{\mL}_n,\bar o_n)$ from its prototype logits through Eq.~(\ref{eq:blend}).

Each feature is weighted by evaluating its Gaussian at the query position. With displacement $\vd_n=\vx-\vx_n$, the Mahalanobis distance is
\begin{equation}
\vu_n=\bar{\mL}_n^{-1}\vd_n,\qquad
m_n=\|\vu_n\|^2
=\vd_n^{\!\top}\bar{\mSigma}_n^{-1}\vd_n,
\end{equation}
where the triangular system $\bar{\mL}_n\vu_n=\vd_n$ is solved directly without explicitly computing $\bar{\mSigma}_n^{-1}$. The weights are normalised across the 27 neighbours to obtain the context vector
\begin{equation}
\begin{gathered}
w_n=\bar o_n\exp\left(-\tfrac12m_n\right),\qquad
\tilde w_n=\frac{w_n}{\sum_{n'}w_{n'}+10^{-8}},\qquad \\
\vgamma(\vx)=\sum_{n=1}^{27}\tilde w_n\vf_n\in\R^8
\end{gathered}
\label{eq:gather}
\end{equation}
Since $\vd_n$ depends on the continuous query coordinate, the gather weights vary within each LR voxel, providing spatially adaptive aggregation of neighbouring features. The frequency branch (Section~\ref{sec:decoder}) additionally encodes the sub-voxel query position and target resolution. At volume boundaries, out-of-bound anchors are clamped to the nearest boundary voxel.

subsection{Implicit decoder}
\label{sec:decoder}

The implicit decoder extends the local texture estimator (LTE) \citep{lee2022lte} to 3D and additionally conditions the prediction on the target voxel size. For a query coordinate $\vx$, we obtain the coefficient and frequency vectors by trilinearly interpolating the corresponding maps from the eight surrounding LR grid points. At each neighbouring point $\vx_n$, the coefficient and frequency vectors are
\begin{equation}
\va_n=\mA(\vx_n)\in\R^{128},\qquad
\vomega_n=\mOmega(\vx_n)\in\R^{192}
\end{equation}
Let $\lambda_n(\vx)$ denote the trilinear interpolation weight of neighbour $n$. The coefficient and frequency vectors at the query position are then
\begin{equation}
\begin{gathered}
\va_{\vx}=\sum_{n=1}^{8}\lambda_n(\vx)\va_n\in\R^{128},\qquad \\
\vomega_{\vx}
=\operatorname{reshape}\left(
\sum_{n=1}^{8}\lambda_n(\vx)\vomega_n,\,64,\,3
\right)
\in\R^{64\times3}
\end{gathered}
\end{equation}
where the 192 interpolated frequency values are reshaped into 64 three-dimensional frequency vectors. We define the position of the query relative to its nearest LR voxel centre $\vx^\ast$ as
\begin{equation}
\vdelta_{\vx}=(\vx-\vx^\ast)\odot(H,W,D)
\end{equation}
The target voxel size is similarly expressed relative to the LR grid,
\begin{equation}
\tilde\cell=\cell\odot(H,W,D)
=\left(\tfrac{2H}{H'},\tfrac{2W}{W'},\tfrac{2D}{D'}\right) \approx \left(\tfrac{2}{s_h},\tfrac{2}{s_w},\tfrac{2}{s_d}\right)
\end{equation}
where $s_h$, $s_w$, and $s_d$ are the upscaling factors along the three axes. A bias-free linear layer maps $\tilde\cell$ to a phase term $\vphi=\mP\tilde\cell\in\R^{64}$. The local Fourier encoding is then
\begin{equation}
\vtheta_{\vx}=\vomega_{\vx}\vdelta_{\vx}+\vphi,\qquad
\ve_{\vx}=
\big[\cos(\pi\vtheta_{\vx});\sin(\pi\vtheta_{\vx})\big]
\in\R^{128} 
\end{equation}

The coefficient vector $\va_{\vx}$ modulates the Fourier encoding element-wise, following LTE. We incorporate the local context from the Gaussian gather by projecting $\vgamma(\vx)$ to the same feature dimension and adding it to this representation:
\begin{equation}
\vv_{\vx}
=\va_{\vx}\odot\ve_{\vx}
+\mW_c\vgamma(\vx)
\qquad
\mW_c:\R^8\rightarrow\R^{128}
\label{eq:fusion}
\end{equation}
Where $\mW_c$ is a linear trainable layer. A four-layer MLP with GELU activation function maps the resulting feature vector to the predicted intensity correction,
\begin{equation}
\mu(\vx)=\MLP(\vv_{\vx})
\end{equation}

Both the relative coordinate $\vdelta_{\vx}$ and the target cell $\tilde\cell$ are provided explicitly to the decoder. The model can therefore be queried at arbitrary spatial coordinates and target voxel sizes without changing its parameters. In particular, specifying the cell independently along each axis allows the same decoder to handle both isotropic and anisotropic upscaling.

\subsection{Residual connection}
\label{sec:residual}

The decoder predicts an intensity correction relative to the trilinearly interpolated LR input. Since the encoder input is in the MAISI range whereas the reconstruction target is in $[0,1]$, we first map the interpolated intensity to the target range,
\begin{equation}
\lrf_{\mathrm{gt}}(\vx)=0.5\,\lrf(\vx)+0.5
\end{equation}
A learned gate controls its contribution at each query location,
\begin{equation}
g(\vx)=\sigmoid\big(G(\vv_{\vx})\big),\qquad
p(\vx)=\mu(\vx)+g(\vx)\,\lrf_{\mathrm{gt}}(\vx)
\end{equation}
followed by a refinement head,
\begin{equation}
\hat y(\vx)=p(\vx)+r\big([p(\vx);\vv_{\vx}]\big)
\end{equation}
The gating head is initialised with zero final-layer weights and a bias of $3.0$ ($g(\vx)\approx0.95$), keeping the initial prediction close to trilinear interpolation. The refinement head is zero-initialised and both components are subsequently learned jointly with the decoder.

\subsection{Measurement consistency}
\label{sec:dc}

Let $\gH_s$ denote the assumed linear degradation operator mapping an HR volume to its LR observation. After reconstruction, we enforce consistency with $\mX$ by projecting $\hat{\mY}$ onto the affine set $\{\mY:\gH_s\mY=\mX\}$:
\begin{equation}
\hat{\mY}\leftarrow\hat{\mY}+\gH_s^{+}\big(\mX-\gH_s\hat{\mY}\big),\qquad
\gH_s^{+}=\gH_s^{\!\top}(\gH_s\gH_s^{\!\top})^{-1}
\label{eq:dc}
\end{equation}
This is the the Euclidean projection under the assumed degradation operator and is applied only at inference without additional training. In practice, $(\gH_s\gH_s^\top)^{-1}$ is computed without explicit matrix inversion by solving the corresponding linear system using conjugate gradients.

\subsection{Training and inference}
\label{sec:training}

We train separate GSMR3D models for contrast-enhanced T1-weighted (CT1) and T1-weighted (T1) MRI. The CT1 development set comprises $3{,}731$ training and validation volumes from the CENTRIC and CORE glioblastoma trials \citep{stupp2014cilengitide,nabors2015two}, of which $3{,}545$ are used for training and the rest for validation. The T1 development set comprises $30{,}254$ training and validation volumes, predominantly from the German National Cohort (NAKO) \citep{peters_FrameworkBaseline_2022}, together with 102 volumes from CENTRIC and CORE, of which $28{,}742$ are used for training and the rest for validation. EORTC-26101 \citep{wick2017lomustine} is excluded entirely from training and validation and serves as an independent tumour test cohort for both contrasts. Full dataset composition is provided in Appendix~\ref{app:data}.

During training, the upscaling factor is sampled independently along each spatial axis as $s_a=1+3u_a^{0.5}$, where $u_a\sim\mathcal U(0,1)$. This exposes the model to a continuous range of isotropic and anisotropic scale combinations in $[1,4]$. HR crops are degraded to fixed $32^3$ LR patches, from which $32{,}768$ HR query points are sampled. The primary objective is an $L_1$ reconstruction loss, supplemented every five iterations with 3D-SSIM, frequency, and edge losses. Models are trained with Adam for 600 epochs, and the checkpoint is selected based on validation loss. During inference, each HR query coordinate gathers from a fixed $3{\times}3{\times}3$ LR neighbourhood, giving a computational complexity of $O(\text{LR encode})+O(\#\text{HR voxels})$. After reconstruction, measurement consistency is enforced using Eq.~(\ref{eq:dc}). Full training and implementation details are provided in Appendix~\ref{app:method}.

\section{Experiments}
\label{sec:experiments}

\paragraph{Setup.} We evaluate GSMR3D on two MRI contrasts and four degradation geometries. The primary evaluation uses 100 tumour cases from EORTC-26101 for both CT1 and T1, with an additional 100 held-out healthy subjects from NAKO dataset for T1. All evaluation cases are independent of the training and validation data. Volumes are resampled to 1\,mm isotropic resolution and used as the HR reference. LR inputs are generated by Gaussian anti-aliasing followed by subsampling for isotropic $2\times$, in-plane $2\times$, through-plane $3\times$, and through-plane $4\times$ SR. A single GSMR3D checkpoint per contrast is evaluated across all four geometries. To assess generalisation to a different tumour pathology, we additionally evaluate the models on 100 brain-metastasis cases from MR-RATE \citep{mrrate}, comprising 670 enhancing metastases lesions. This cohort is also independent of the training and validation data. Further details of the datasets and evaluation protocol are provided in Appendices~\ref{app:data} and~\ref{app:bm-seg}.

We compare GSMR3D with arbitrary-scale methods, geometry-specific specialists, and conventional baselines. The arbitrary-scale methods are an NCSN++ \citep{song2021scorebased} diffusion prior with DDNM-style data consistency \citep{wang2023zeroshot}, ResCNN \citep{du2020super}, and total-variation (TV) reconstruction. The geometry-specific specialists are ArSSR \citep{wu2023arssr} for isotropic, LTE \citep{lee2022lte} for in-plane, and SAINT \citep{peng2020saint} for through-plane SR. We additionally evaluate cubic interpolation and SynthSR \citep{iglesias2023synthsr}. Baseline implementation and training details are provided in Appendix~\ref{app:baseline_training}. We report PSNR and SSIM over the full 3D volume, together with LPIPS \citep{zhang:lpips}, DISTS \citep{ding:dists}, HaarPSI \citep{reisenhofer:haarpsi}, and FID \citep{heusel:ganstrained}. Metric and evaluation details are provided in Appendix~\ref{app:metrics}.

\subsection{Image quality}
\label{sec:main}

\begin{table*}[tp]
\centering
\caption{\textbf{Contrast-enhanced T1 (brain-tumour cohort, $n=100$).}
Specialists are ArSSR, LTE, and SAINT for isotropic, in-plane, and
through-plane SR, respectively. Best in bold; additional results are in
Appendix~\ref{app:results} (Table~\ref{tab:ct1-full}).}
\label{tab:ct1}
\scriptsize
\setlength{\tabcolsep}{2.6pt}
\begin{tabular}{l ccc ccc ccc ccc}
\toprule
 & \multicolumn{3}{c}{Isotropic $2\times$} & \multicolumn{3}{c}{In-plane $2\times$} & \multicolumn{3}{c}{Through-plane $3\times$} & \multicolumn{3}{c}{Through-plane $4\times$} \\
\cmidrule(lr){2-4} \cmidrule(lr){5-7} \cmidrule(lr){8-10} \cmidrule(lr){11-13}
Method & PSNR$\uparrow$ & LPIPS$\downarrow$ & FID$\downarrow$ & PSNR$\uparrow$ & LPIPS$\downarrow$ & FID$\downarrow$ & PSNR$\uparrow$ & LPIPS$\downarrow$ & FID$\downarrow$ & PSNR$\uparrow$ & LPIPS$\downarrow$ & FID$\downarrow$ \\
\midrule
GSMR3D (ours) & \textbf{32.50} & \textbf{0.029} & \textbf{5.6} & 32.69 & \textbf{0.020} & \textbf{3.7} & 32.95 & \textbf{0.025} & \textbf{7.2} & 30.86 & \textbf{0.045} & \textbf{10.8} \\
Diffusion + DC & 31.84 & 0.045 & 6.3 & 32.89 & 0.031 & 4.4 & 33.14 & 0.038 & 10.3 & 30.69 & 0.060 & 12.8 \\
Specialist & 32.24 & 0.043 & 9.4 & 33.72 & 0.027 & 4.1 & \textbf{34.36} & 0.036 & 11.2 & \textbf{31.92} & 0.059 & 16.2 \\
Cubic & 26.98 & 0.147 & 20.8 & 28.22 & 0.106 & 15.2 & 29.12 & 0.118 & 23.3 & 27.33 & 0.162 & 31.4 \\
ResCNN & 32.47 & 0.045 & 8.2 & \textbf{33.73} & 0.029 & 5.3 & 33.50 & 0.038 & 12.7 & 31.15 & 0.061 & 17.0 \\
\bottomrule
\end{tabular}
\end{table*}

Table~\ref{tab:ct1} reports the performance of GSMR3D on CT1 scans. The corresponding CT1 results for all six image quality metrics and the T1 results are provided in Appendix~\ref{app:results} (Tables~\ref{tab:ct1-full} and~\ref{tab:t1}). At isotropic $2\times$ SR, GSMR3D performs best across the evaluated image quality metrics on both CT1 and T1, despite using the same checkpoint across all degradation geometries. Compared with the isotropic specialist ArSSR, PSNR3D increases by 0.3\,dB on CT1 and 1.1\,dB on T1, while FID decreases from 9.4 to 5.6 and from 8.8 to 5.3, respectively. GSMR3D also outperforms the diffusion prior and ResCNN across all metrics in this setting. These results show that using a single model across acquisition geometries does not compromise performance relative to the isotropic specialist in this setting.

For anisotropic SR, the geometry-specific specialists achieve higher distortion-based performance, with approximately 0.4--1.4\,dB higher PSNR3D and generally higher SSIM3D and HaarPSI. In contrast, GSMR3D consistently achieves lower LPIPS, DISTS, and FID. This trade-off is particularly apparent for thick-slice reconstruction. On CT1 through-plane $3\times$ and $4\times$, GSMR3D achieves FID values of 7.2 and 10.8 compared with 11.2 and 16.2 for SAINT. The same overall pattern is observed on T1 (Appendix~\ref{app:results}). Qualitative examples further show sharper reconstructions with GSMR3D, although subtle block-like artefacts are visible along the through-plane direction (Appendix~\ref{app:qual}, Figs.~\ref{fig:app-qual-ct1} and~\ref{fig:app-qual-t1}). These results are consistent with the well-known distortion--perception trade-off, where improvements in distortion-based measures such as PSNR do not necessarily correspond to improvements in perceptual quality. This motivates the downstream anatomical evaluation in Section~\ref{sec:downstream}.

The diffusion prior generally lies between GSMR3D and the geometry-specific specialists in distortion-based performance. In the anisotropic settings, it achieves approximately 0.2--1.1\,dB higher PSNR3D than GSMR3D across both primary test cohorts, while GSMR3D generally achieves lower LPIPS and DISTS. GSMR3D is also substantially faster than diffusion-based sampling, as discussed in Section~\ref{sec:efficiency}. Complete comparisons with TV, SynthSR, cubic interpolation, and ResCNN are reported in Appendix~\ref{app:results}.

We further evaluate image quality on the independent brain-metastasis cohort from MR-RATE, which was not used for training or validation. The complete CT1 and T1 results are reported in Tables~\ref{tab:bm-ct1} and~\ref{tab:bm-t1} in Appendix~\ref{app:results}. On CT1, GSMR3D achieves the best performance across all six metrics at isotropic $2\times$. For anisotropic reconstruction, the same trade-off observed on the primary test cohorts persists. The geometry-specific specialists generally achieve higher distortion-based performance, whereas GSMR3D performs better on the perceptual metrics. At through-plane $3\times$, for example, SAINT achieves higher PSNR (34.30 vs.\ 33.37\,dB), while GSMR3D reduces FID from 10.5 to 4.9 and LPIPS from 0.031 to 0.020. At $4\times$, GSMR3D similarly obtains an FID of 9.1 compared with 15.7 for SAINT. The pattern is consistent on T1. GSMR3D again performs best across all six metrics at isotropic $2\times$, while at in-plane $2\times$ its PSNR is within 0.1\,dB of LTE while achieving better LPIPS, DISTS, HaarPSI, and FID. For through-plane reconstruction, the specialist retains higher distortion-based performance, whereas GSMR3D achieves better perceptual scores at both $3\times$ and $4\times$. These results show that the reconstruction behaviour observed on the primary test cohorts generalises to an external dataset containing brain metastases, without retraining or fine-tuning.

\subsection{Downstream anatomical fidelity}
\label{sec:downstream}

Improvements in image-quality metrics do not necessarily translate into better preservation of clinically relevant information. We therefore evaluate whether SR volumes retain anatomical information required by downstream imaging tasks. For tumour scans, we apply HD-GLIO \citep{kickingereder2019automated} to the held-out glioma cohort and measure segmentation agreement between SR and true HR volumes. We additionally evaluate brain-metastasis segmentation using HD-BM \citep{hd-bm} on an independent cohort of 100 patients, assessing both segmentation agreement and lesion detection. For T1, we perform 20-structure brain parcellation using SynthSeg \citep{billot2023synthseg} on 100 healthy NAKO subjects. Together, these experiments assess whether reconstructed images preserve anatomical and pathological information across different populations and downstream tasks.

\begin{table*}[tp]
\centering
\scriptsize
\setlength{\tabcolsep}{2pt}
\begin{minipage}[t]{0.49\linewidth}
\centering
\caption{\textbf{Downstream tumour segmentation} (CT1, $n{=}100$). Dice agreement between HD-GLIO run on the SR and on the true HR volume, contrast-enhancing (CE) and whole tumour (WT). Full results in Appendix~\ref{app:seg}.}
\label{tab:seg}
\adjustbox{max width=\linewidth}{%
\begin{tabular}{lcccccccc}
\toprule
 & \multicolumn{2}{c}{Iso $2\times$} & \multicolumn{2}{c}{In-pl.\ $2\times$} & \multicolumn{2}{c}{Th.\ $3\times$} & \multicolumn{2}{c}{Th.\ $4\times$} \\
\cmidrule(lr){2-3} \cmidrule(lr){4-5} \cmidrule(lr){6-7} \cmidrule(lr){8-9}
Method & CE & WT & CE & WT & CE & WT & CE & WT \\
\midrule
GSMR3D (ours) & \textbf{0.961} & \textbf{0.994} & \textbf{0.967} & \textbf{0.995} & \textbf{0.947} & \textbf{0.992} & \textbf{0.905} & \textbf{0.986} \\
Diffusion + DC & 0.946 & 0.992 & 0.956 & 0.993 & 0.928 & 0.990 & 0.886 & 0.983 \\
Specialist & 0.955 & 0.992 & 0.966 & 0.994 & 0.933 & 0.991 & 0.900 & 0.985 \\
TV & 0.941 & 0.989 & 0.951 & 0.991 & 0.914 & 0.988 & 0.875 & 0.980 \\
Cubic & 0.859 & 0.968 & 0.901 & 0.974 & 0.870 & 0.977 & 0.824 & 0.967 \\
ResCNN & 0.949 & 0.992 & 0.966 & 0.994 & 0.927 & 0.991 & 0.901 & 0.984 \\
\bottomrule
\end{tabular}}
\end{minipage}
\hfill
\begin{minipage}[t]{0.49\linewidth}
\centering
\caption{\textbf{Downstream brain parcellation} (T1, NAKO, $n{=}100$). Dice agreement with the parcellation of the true HR volume, mean over 20 structures (All) and cerebral cortex. Per-structure results in Appendix~\ref{app:parc}.}
\label{tab:parc}
\adjustbox{max width=\linewidth}{%
\begin{tabular}{lcccccccc}
\toprule
 & \multicolumn{2}{c}{Iso $2\times$} & \multicolumn{2}{c}{In-pl.\ $2\times$} & \multicolumn{2}{c}{Th.\ $3\times$} & \multicolumn{2}{c}{Th.\ $4\times$} \\
\cmidrule(lr){2-3} \cmidrule(lr){4-5} \cmidrule(lr){6-7} \cmidrule(lr){8-9}
Method & All & Cortex & All & Cortex & All & Cortex & All & Cortex \\
\midrule
GSMR3D (ours) & \textbf{0.940} & \textbf{0.953} & \textbf{0.956} & \textbf{0.962} & 0.940 & 0.943 & 0.923 & 0.921 \\
Diffusion + DC & 0.932 & 0.939 & 0.951 & 0.955 & 0.938 & 0.941 & 0.921 & 0.915 \\
Specialist & 0.931 & 0.947 & 0.954 & 0.961 & \textbf{0.943} & \textbf{0.954} & \textbf{0.928} & \textbf{0.936} \\
TV & 0.919 & 0.910 & 0.941 & 0.938 & 0.922 & 0.914 & 0.901 & 0.877 \\
Cubic & 0.915 & 0.903 & 0.931 & 0.922 & 0.919 & 0.907 & 0.901 & 0.875 \\
ResCNN & 0.930 & 0.944 & 0.950 & 0.958 & 0.942 & 0.953 & 0.926 & 0.936 \\
\bottomrule
\end{tabular}}
\end{minipage}
\end{table*}


For tumour segmentation, GSMR3D achieves the highest agreement with the HR-derived segmentation across all four geometries for both contrast-enhancing (CE) and whole-tumour (WT) regions (Table~\ref{tab:seg}). This is particularly notable for through-plane SR, where SAINT achieves higher PSNR3D. At through-plane $3\times$, GSMR3D achieves a CE agreement of 0.947, compared with 0.933 for SAINT, 0.928 for the diffusion prior, and 0.927 for ResCNN. Paired Wilcoxon tests on CE Dice give $p<10^{-4}$ against all three methods in this setting. Across all four geometries, GSMR3D exceeds ResCNN, the corresponding specialist, and the diffusion prior in all twelve paired comparisons, with $p\leq0.03$ throughout (Appendix~\ref{app:seg}). Thus, the higher PSNR3D achieved by the through-plane specialist does not translate into closer agreement with the segmentation obtained from the true HR image. Representative tumour reconstructions and corresponding HD-GLIO segmentations are shown in Fig.~\ref{fig:app-qual-tumour} in Appendix~\ref{app:qual}.

The same conclusion is supported by the extended tumour analysis against expert annotations. Differences between methods are smaller in this comparison because the segmentation obtained from the true HR volume is itself imperfect relative to the expert labels. Nevertheless, the super-resolved inputs remain close to the HR reference, and the full results in Appendix~\ref{app:seg} show that GSMR3D combines high agreement with the HR-derived segmentation with low tumour-volume error. 

Brain parcellation shows a greater dependence on the degradation geometry and the super-resolution model (Table~\ref{tab:parc}). GSMR3D achieves the highest mean Dice agreement at isotropic and in-plane $2\times$, whereas the geometry-specific specialist and ResCNN perform slightly better at through-plane $3\times$ and $4\times$. Per-structure analysis shows that these differences are primarily concentrated in the cerebral cortex, cerebral white matter, CSF, and lateral ventricles. For several subcortical structures and the brainstem, GSMR3D instead performs best or remains comparable to the leading method. Across all four geometries, GSMR3D consistently achieves higher mean parcellation agreement than the diffusion prior. Detailed per-structure results are provided in Appendix~\ref{app:parc}.

We additionally assess generalisation to a different tumour pathology using an independent cohort of 100 brain-metastasis cases containing 670 enhancing lesions. Brain metastases are generally small lesions and therefore provide a relevant test of whether SR preserves pathological structures that may be obscured at lower spatial resolution. HD-BM \citep{hd-bm} is applied to the reconstructed volumes to evaluate segmentation agreement and lesion detection relative to the original HR images. GSMR3D achieves the highest lesion sensitivity across all four geometries, recovering 99.4\% and 99.6\% of lesions at isotropic and in-plane $2\times$, respectively. The advantage persists under substantial through-plane degradation. At through-plane $4\times$, GSMR3D retains 91.5\% of lesions, compared with 90.7\% for SAINT, 87.3\% for ResCNN, and 86.7\% for the diffusion prior. This is illustrated in a representative case containing 70 HR-derived lesions (Fig.~\ref{fig:app-qual-bm}), where GSMR3D detects all 70 lesions at through-plane $3\times$ and 69 at $4\times$. At $4\times$, SAINT detects 64 lesions, the diffusion prior 62, and ResCNN 59. GSMR3D also achieves the highest CE segmentation agreement at isotropic $2\times$ and both through-plane geometries. These results show that GSMR3D can preserve small focal lesions that are particularly sensitive to loss of spatial resolution. Detailed quantitative results are provided in Appendix~\ref{app:bm-seg}.

Overall, the downstream experiments show that conventional image-quality metrics do not fully capture anatomical fidelity. GSMR3D best preserves tumour-related information, achieving the highest glioma segmentation agreement and brain-metastasis lesion sensitivity across all geometries, including settings in which specialists achieve higher PSNR3D. Brain parcellation is more geometry dependent, with specialists and ResCNN performing better for through-plane reconstruction.

\subsection{Computational efficiency}
\label{sec:efficiency}

GSMR3D encodes the LR volume once and evaluates each HR query over a fixed LR neighbourhood. The computation per query is therefore independent of the upscaling factor, while total runtime scales with the number of reconstructed HR voxels. Across the four evaluation geometries, reconstruction takes 5.8--6.4\,s on a single NVIDIA A100 GPU (Table~\ref{tab:runtime}), compared with 137--149\,s for the diffusion prior and approximately 31\,s for TV. GSMR3D is therefore approximately 13 times faster than the diffusion baseline. Geometry-specific specialists are slightly faster in some settings but require separate models for different geometries, whereas a single GSMR3D checkpoint per contrast handles all four. SynthSR was evaluated on CPU because of GPU incompatibility and is therefore not directly comparable. Full runtime results are provided in Appendix~\ref{app:results}.

\section{Conclusion}

\label{sec:conclusion}

Our proposed GSMR3D provides a single framework for arbitrary-scale isotropic and anisotropic 3D MRI super-resolution, combining learned anisotropic Gaussian prototypes with continuous query-based reconstruction and measurement consistency. Across T1-weighted and contrast-enhanced T1-weighted brain MRI, a single checkpoint per contrast performs competitively with geometry-specific specialists across different acquisition geometries and outperforms the isotropic specialist. Although geometry-specific models achieve higher PSNR in some anisotropic settings, GSMR3D consistently achieves better perceptual metrices. More importantly, GSMR3D most faithfully preserves glioma segmentations across all evaluated geometries and achieves the highest lesion sensitivity on an independent brain-metastasis cohort, demonstrating that differences in conventional image-quality metrics do not necessarily reflect preservation of clinically relevant information. Brain parcellation is more geometry dependent, with GSMR3D performing best for isotropic and in-plane reconstruction while specialist models perform better in some through-plane settings. GSMR3D also reconstructs a volume in approximately 6\,s, which is approximately $13\times$ faster than the evaluated diffusion approach. Together, these results demonstrate the potential of a single model to perform MRI super-resolution across different spatial resolutions and acquisition geometries without geometry-specific retraining. With prospective validation on accelerated acquisitions, this approach could enable shorter MRI examinations while preserving diagnostically relevant information, with the potential to reduce scan times and improve patient throughput.

\subsection{Limitations}

The current evaluation is restricted to brain MRI and synthetically generated LR inputs. Although the degradation model enables controlled evaluation across multiple acquisition geometries and acceleration factors, it does not capture all characteristics of prospectively accelerated MRI, including acquisition-specific noise, motion, and other scanner-dependent effects. Evaluation on prospectively acquired LR data is therefore needed to establish how well the observed reconstruction and pathology-preservation performance translates to clinical acquisitions. The experiments are also limited to T1-weighted and contrast-enhanced T1-weighted MRI. Further evaluation across additional MRI contrasts, scanners, acquisition protocols, anatomies, and pathologies will be important for establishing generalisability and determining clinically appropriate acceleration factors.

\section{Reproducibility statement}
We provide the methodological and experimental details required to reproduce the proposed method and the experiments reported in this work. The architecture and methodological details are described in Section~\ref{sec:method} and Appendix~\ref{app:method}. These include the construction of the Gaussian dictionary, anisotropic Gaussian parameterization, the frequency branch, query-based reconstruction, and the implementation of the fused CUDA operation. We additionally specify the optimization and training procedure, loss functions and their weights, learning-rate schedule, sampling strategy, and other hyperparameters required to reproduce the reported models.

The datasets, preprocessing procedures, train/validation/test splits, and evaluation protocol are described in Section~\ref{sec:experiments} and the corresponding appendices. We report the evaluation metrics and experimental settings used for all quantitative comparisons to facilitate reproduction of the reported results.

The implementation, including the training and inference code and the fused CUDA kernel, will be made publicly available upon publication.

\section{AI Use Statement}

A generative AI tool, namely ChatGPT 5.6 sol from OpenAI, was used only for minor edits to the manuscript text to improve readability, clarity, and conciseness.

\section{Conflict of Interest}

PV reports research funding from the German Research Foundation (DFG), the Else Kröner Fresenius Foundation, and the European Research Council (ERC), as well as consulting fees and stock options from Need Inc., all outside the submitted work. GB reports consulting fees from Cercare Medical outside the submitted work. FB reports an unrestricted research grant from Siemens Healthineers and speaker and consultancy fees from Bayer Healthcare, all outside the submitted work. All other authors declare that they have no competing interests. 

\section{Acknowledgements}

We acknowledge funding from the European Research Council (ERC Consolidator Grant AI-NEXT, project ID: 101171462), the German Research Foundation (DFG) as part of the Priority Programme 2177 “Radiomics: Next Generation of Biomedical Imaging” (project ID: 428223917) and Collaborative Research Center 1389 (UNITE Glioblastoma – project ID: 404521405). PV is partially funded by an Else Kröner Clinician Scientist Endowed Professorship from the Else Kröner Fresenius Foundation (Ref. No.: 2022\_EKCS.17). AR is partially funded by the Faculty of Medicine at University Hospital Bonn through the BONFOR Startup Postdoc Fellowship (Ref. No.: 2024-1B-10)

Parts of the MRI data used in this study were obtained from the German National Cohort (NAKO; Application No. NAKO-999). NAKO is funded by the Federal Ministry of Research, Technology and Space (BMFTR; project funding Ref. Nos. 01ER1301A/B/C, 01ER1511D, 01ER1801A/B/C/D, and 01ER2301A/B/C), the Federal States of Germany, the Helmholtz Association, the participating universities, and the institutes of the Leibniz Association. We thank all NAKO participants and the staff of this research initiative. Members and affiliations of the NAKO Investigator Consortium are available at \url{https://www.nako.de/principal-investigators}.

Computational resources were provided by the University Hospital Bonn (UKB) and the German Cancer Research Center (DKFZ). CT1 model training and evaluation were performed on NVIDIA A100 GPUs at DKFZ. For T1, initial training on NAKO data was performed on NVIDIA A100 GPUs at UKB, while subsequent fine-tuning on CENTRIC and CORE and all evaluations were performed at DKFZ.


\bibliographystyle{cas-model2-names}

\bibliography{references,litreview,gsreview}

\appendix
\clearpage
\begin{center}
  {\Large\bfseries Appendix}\\[4pt]
  {\normalsize Arbitrary-Scale Anisotropic 3D MRI Super-Resolution with Gaussian Prototypes}
\end{center}
\vspace{6pt}
\noindent The appendix is organised as follows.
Appendix~\ref{app:related} extends the related work.
Appendix~\ref{app:method} gives the training data pipeline, losses, inference procedure and the full flowchart (Figure~\ref{fig:details}). It also describes how every baseline was trained.
Appendix~\ref{app:results} reports the full metric tables, runtime, downstream results, qualitative figures and ablations.
\setcounter{figure}{0}
\renewcommand{\thefigure}{A\arabic{figure}}
\setcounter{table}{0}
\renewcommand{\thetable}{A\arabic{table}}
\renewcommand{\theHfigure}{A\arabic{figure}}
\renewcommand{\theHtable}{A\arabic{table}}

\section{Extended related work}
\label{app:related}

\paragraph{Gaussian representations in medical imaging.}
Gaussian representations have recently been explored for several medical reconstruction problems. 3DGSMR fits complex-valued 3D Gaussians directly to undersampled $k$-space through self-supervised, per-scan optimisation \citep{peng2025three}. CauGD$^2$-Net incorporates Gaussian splatting into a dual-domain reconstruction network \citep{houCausalitydrivenDualdomainNetwork2026}, while compressed 3DGS has been investigated for efficient rendering of volumetric anatomy \citep{niedermayrApplication3DGaussian2024}. For fetal slice-to-volume reconstruction, \citet{danneckerFastExplicitSlicetoVolume} represent the HR volume with anisotropic 3D Gaussians and exploit their closure under convolution to model the acquisition point-spread function analytically as $\mSigma_{\mathrm{obs}}=\mSigma_{\mathrm{HR}}+\mSigma_{\mathrm{PSF}}$.

Most closely related to our setting, the concurrent PhyMRI-SR \citep{phymrisr2026} formulates 2D MRI SR using Gaussian splatting. It derives a fixed covariance dictionary from Gaussians fitted to HR images and incorporates a T2-weighted signal model, predicting proton density and an effective transverse relaxation term rather than image intensity directly. GSMR3D instead jointly learns a volumetric dictionary of anisotropic Gaussian prototypes without assuming a sequence-specific signal model, allowing the same formulation to be applied to both T1 and contrast-enhanced T1 MRI. PhyMRI-SR scatters Gaussians onto the 2D target grid, whereas GSMR3D gathers their contributions from a fixed 3D LR neighbourhood for each HR query. This provides continuous Gaussian weighting in three dimensions while keeping the number of candidate contributions per query fixed. The two approaches also incorporate acquisition information differently: PhyMRI-SR embeds an explicit MRI signal model, while GSMR3D applies a measurement projection under the assumed degradation operator.

\paragraph{Generative priors, data consistency, and faithfulness.}
Diffusion and flow models have increasingly been applied to medical image reconstruction and SR \citep{khateriMRISuperResolutionDeep2025}. Applications in MRI include single-contrast SR \citep{wuSuperresolutionBrainMRI2023}, multi-contrast SR \citep{mao2023disc,li2024rethinking}, joint denoising and SR \citep{chungMRImageDenoising2023}, spectroscopic imaging \citep{dongFlowbasedTruncatedDenoising2025}, and 3D synthesis in the wavelet domain \citep{danese_FlowLetConditional_2026}. Generative priors can also be combined with an assumed forward model to solve inverse problems through iterative sampling and data-consistency updates \citep{song2022solving,hu2024restoration,lee2023improving,luo2024autoregressive}.

These approaches provide strong generative priors but introduce two concerns for medical SR. First, iterative sampling increases reconstruction time relative to feed-forward methods. Second, plausible high-frequency detail need not correspond to the underlying anatomy, and reconstruction errors can affect anatomical structures and derived measurements \citep{kim_MPFlowMultimodal_2026,morell-ortegaCAHALClinicallyApplicable2026}. GSMR3D therefore uses deterministic reconstruction followed by a single closed-form projection that enforces consistency with the LR measurements under the assumed degradation operator. We additionally reuse the frozen full-resolution stem of the pretrained 3D MAISI encoder \citep{guo2024maisi,zhao2026maisi}, preserving an LR-grid feature representation from which the Gaussian prototypes are selected.

\section{Method details}
\label{app:method}

\begin{figure*}[t]
  \centering
  \includegraphics[width=\linewidth]{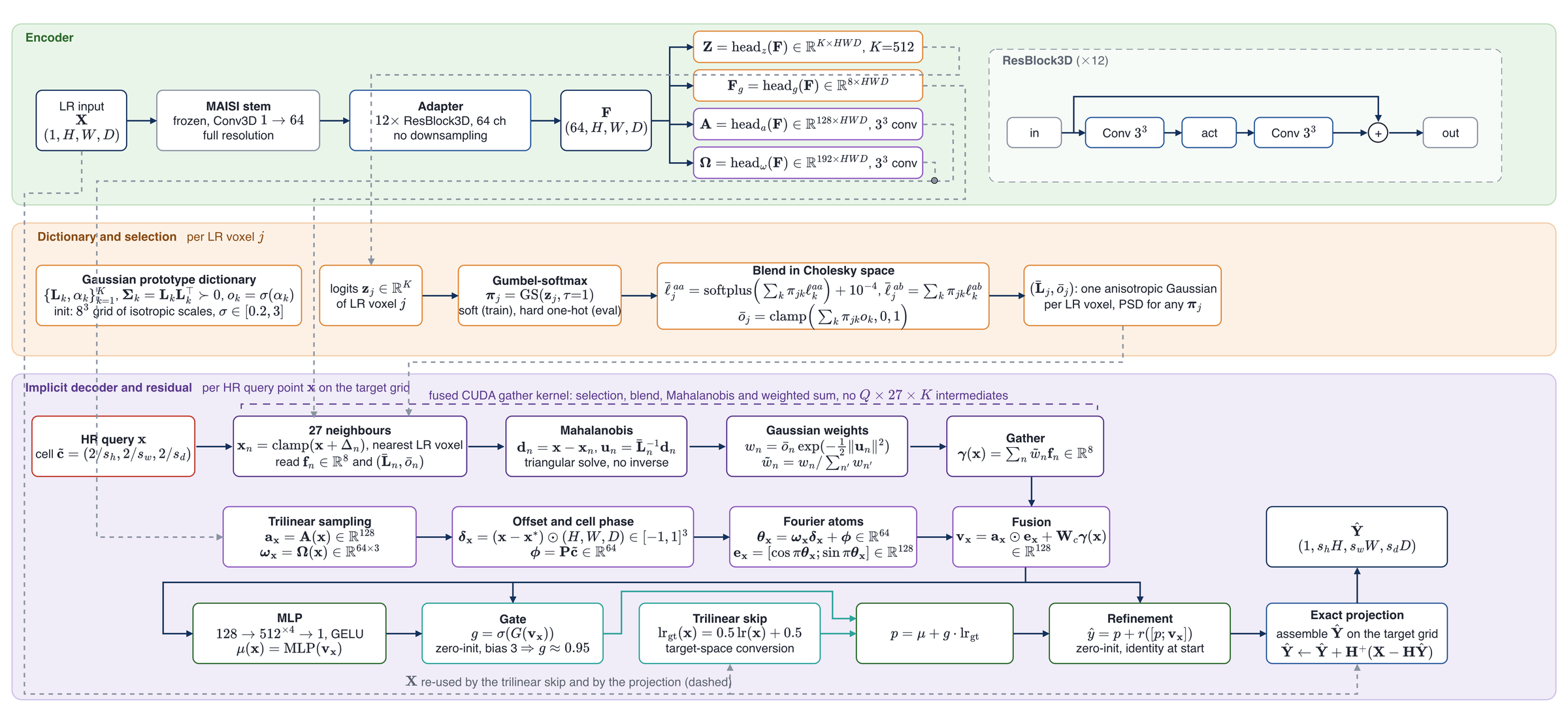}
 \caption{\textbf{Detailed GSMR3D architecture.} Top: the LR volume is processed by a frozen MAISI stem and a trainable residual adapter, preserving the LR grid while producing the feature volume $\mF$. Separate heads predict prototype logits $\mZ$, gather features $\mF_g$, and coefficient and frequency maps $\mA$ and $\mOmega$. Middle: at each LR voxel, the prototype logits select and blend elements of a shared dictionary of $K{=}512$ anisotropic 3D Gaussians in Cholesky space, yielding a local Gaussian with positive-definite covariance. Bottom: for each HR query coordinate $\vx$, features from a fixed $3{\times}3{\times}3$ LR neighbourhood are weighted by the selected Gaussians and gathered into $\vgamma(\vx)$. The sampled coefficient and frequency maps form scale-conditioned Fourier features, which are fused with $\vgamma(\vx)$ and decoded by an MLP. The resulting residual is combined with a gated trilinear LR skip and passed through a zero-initialised refinement head. The reconstructed queries are assembled on the target grid, and a final a projection enforces measurement consistency under the assumed degradation operator. }
  \label{fig:details}
\end{figure*}

\subsection{Training data pipeline}
For each sampled scale, we extract a random HR crop of size $\operatorname{round}(32s_a)$ along each axis, percentile-clip intensities to $[p_1,p_{99.5}]$, map them to $[0,1]$, and downsample the crop to a fixed $32^3$ LR patch using Gaussian blurring and downsampling (the same operator $\gH_s$ used at test time). We uniformly sample $Q=32{,}768$ HR voxels as queries. For the spatial losses, we additionally sample a contiguous $32^3$ HR sub-volume, since these losses require local spatial structure. While training, random flips along all three axes are used for data augmentation. We retain only the first encoder block of the pretrained MAISI VAE because subsequent blocks downsample the feature maps, whereas GSMR3D requires voxel-aligned features at the original LR resolution. Early convolutional features are generally more transferable across tasks \citep{yosinski2014transferable}, and we therefore keep the MAISI stem frozen and learn a residual adapter for task-specific feature extraction.

\paragraph{Training loss}
During training, the primary objective is an $L_1$ loss over the randomly sampled HR queries. At every query point $\vx_q$, the loss is $\gL_1=\frac{1}{Q}\sum_q|\hat{\mY}(\vx_q)-\mY(\vx_q)|$. While this provides direct voxel-wise supervision, $L_1$ alone does not explicitly encourage preservation of local structure, spatial-frequency content, or anatomical edges. To address these limitations, at every fifth training step we additionally predict a contiguous HR sub-volume and reshape the queries into a volume $\hat P$ to compute three spatial losses. Computing these losses over the full HR region corresponding to the $32^3$ LR patch at every step would require approximately $s_hs_ws_d$ times as many queries, while keeping the query count fixed would restrict the losses to a smaller spatial region. We therefore apply these losses every fifth training step. First, we use a 3D SSIM loss, $1-\mathrm{SSIM}(\hat P,P)$, with a $7^3$ Gaussian window ($\sigma=1.5$, $C_1=0.01^2$, $C_2=0.03^2$) to encourage local structural similarity. Second, an FFT-magnitude loss, $\big\||\mathcal F\hat P|-|\mathcal F P|\big\|_1$, encourages preservation of spatial-frequency content. Finally, an edge-weighted $L_1$ loss places greater emphasis on anatomical boundaries. We compute the magnitude of the 3D 7-point Laplacian of the target, $\lambda=|\Delta_{7\mathrm{pt}}P|$, and transform it into weights $\omega=1+\log(1+\lambda)/\max\log(1+\lambda)\in[1,2]$, such that voxels with stronger local intensity transitions receive greater weight. The complete objective is:

\begin{equation}
\gL=\gL_1+\mathbb{1} [t\bmod5=0]\,\big(0.05\,\gL_{\mathrm{SSIM}}+10^{-4}\,\gL_{\mathrm{FFT}}+0.05\,\gL_{\mathrm{edge}}\big)
\end{equation}

For optimisation, we use Adam with an initial learning rate of $10^{-4}$ and a global batch size of 48 volumes across eight A100 GPUs using distributed data parallelism (DDP). The CT1 model is trained for 600 epochs on the CENTRIC and CORE data. The T1 model is first trained for 500 epochs using only NAKO data, with the learning rate halved every 100 epochs. The resulting model is subsequently fine-tuned for 100 epochs on the T1 images from CENTRIC and CORE using $2\times$ the final learning rate of the initial training stage.

\paragraph{Dataset}
\label{app:data}

\begin{table}[!tb]
\centering
\caption{\textbf{Training data.} Number of volumes from each source after filtering and the 95/5 training--validation split. The CT1 model is trained on CENTRIC and CORE. The T1 model is trained for 500 epochs on NAKO and subsequently fine-tuned for 100 epochs on CENTRIC and CORE.}
\label{tab:data}
\scriptsize
\setlength{\tabcolsep}{3pt}
\renewcommand{\arraystretch}{0.9}
\adjustbox{max width=\columnwidth}{%
\begin{tabular}{llrrrr}
\toprule
Contrast & Source & Available & Dropped & Train & Val \\
\midrule
CT1 & CENTRIC $+$ CORE & 3{,}836 & 105 & 3{,}545 & 186 \\
\midrule
\multirow{3}{*}{T1} & NAKO & 30{,}152 & 0 & 28{,}645 & 1{,}507 \\
 & CENTRIC $+$ CORE & 102 & 0 & 97 & 5 \\
 & Total & 30{,}254 & 0 & 28{,}742 & 1{,}512 \\
\bottomrule
\end{tabular}}
\end{table}

Table~\ref{tab:data} summarises the training data. The CT1 model is trained on pre-operative and follow-up scans from the CENTRIC \citep{stupp2014cilengitide} and CORE \citep{nabors2015two} trials of glioblastoma, an aggressive primary brain tumour. The T1 model is initially trained for 500 epochs on healthy subjects from the German National Cohort (NAKO) \citep{peters_FrameworkBaseline_2022} and subsequently fine-tuned for 100 epochs on T1 scans from the same EORTC trials. EORTC-26101 \citep{wick2017lomustine} is excluded entirely from training and validation and serves as the tumour test cohort for both contrasts. Healthy T1 test subjects are drawn from NAKO and are likewise excluded from training and validation. Volumes with any spatial dimension below 128 voxels are excluded ($n=105$), after which each source is split 95/5 at the volume level into training and validation sets.

\subsection{Inference}
\paragraph{Inference}
At inference, GSMR3D accepts an LR volume of arbitrary size and a target grid with any per-axis scaling factors, including non-integer and anisotropic scales. The LR volume is encoded once, after which HR coordinates are evaluated in chunks of 30{,}000 queries, with each query gathering from a fixed $3{\times}3{\times}3$ LR neighbourhood. This chunk size keeps memory usage at approximately 11\,GB for isotropic $2\times$ SR, making inference feasible on GPUs with moderate memory capacity. The chunk size can be increased when additional GPU memory is available without changing the reconstruction. The predictions are reshaped to the target grid, clamped to $[0,1]$, and measurement consistency is enforced using Eq.~(\ref{eq:dc}). The computational cost is $O(\text{LR encode})+O(\#\text{HR voxels})$, with a fixed 27 neighbours evaluated per query. Since the Gaussian gather is repeated for every HR query, we implement it as a fused CUDA kernel to perform the neighbour weighting and feature aggregation efficiently on the GPU. Each reported runtime corresponds to an independent end-to-end reconstruction without caching between SR configurations. For isotropic $2\times$ SR, which produces eight times as many output voxels as the LR input, reconstruction takes approximately 6\,s on a single NVIDIA A100 GPU. When multiple target resolutions are required from the same LR volume, the encoder output can be cached and reused, further reducing the total evaluation time.

\begin{table*}[!tbp]
\centering
\caption{\textbf{Image quality analysis on contrast-enhanced T1 volumes (brain-metastasis cohort, $n{=}100$).} Same protocol as Table~\ref{tab:ct1}}
\label{tab:bm-ct1}
\begin{adjustbox}{max width=0.8\linewidth}
\footnotesize
\renewcommand{\arraystretch}{0.9}
\setlength{\tabcolsep}{5pt}
\begin{tabular}{lcccccc}
\toprule
Method & PSNR3D$\uparrow$ & SSIM3D$\uparrow$ & LPIPS$\downarrow$ & DISTS$\downarrow$ & HaarPSI$\uparrow$ & FID$\downarrow$ \\
\midrule
\multicolumn{7}{l}{\textit{Isotropic $2\times$}} \\
GSMR3D (ours) & \textbf{33.84} & \textbf{0.982} & \textbf{0.022} & \textbf{0.060} & \textbf{0.812} & \textbf{4.5} \\
Diffusion prior + DC & 33.49 & 0.979 & 0.033 & 0.071 & 0.803 & 6.7 \\
ArSSR (specialist) & 33.70 & 0.980 & 0.030 & 0.076 & 0.760 & 8.6 \\
TV & 31.93 & 0.972 & 0.045 & 0.088 & 0.749 & 10.3 \\
Cubic & 28.32 & 0.937 & 0.121 & 0.127 & 0.659 & 20.6 \\
ResCNN model & \textbf{33.71} & 0.982 & 0.032 & 0.074 & 0.802 & 8.2 \\
SynthSR & 21.92 & 0.689 & 0.107 & 0.196 & 0.599 & 53.3 \\
\midrule
\multicolumn{7}{l}{\textit{In-plane $2\times$}} \\
GSMR3D (ours) & 34.98 & \textbf{0.988} & \textbf{0.016} & 0.050 & 0.858 & \textbf{3.0} \\
Diffusion prior + DC & 35.41 & 0.986 & 0.020 & 0.053 & 0.849 & 3.7 \\
LTE (specialist) & \textbf{35.71} & \textbf{0.988} & 0.016 & \textbf{0.049} & \textbf{0.864} & 3.3 \\
TV & 33.60 & 0.981 & 0.026 & 0.065 & 0.806 & 5.3 \\
Cubic & 30.06 & 0.958 & 0.081 & 0.105 & 0.721 & 13.9 \\
ResCNN model & 35.33 & 0.986 & 0.018 & 0.052 & 0.857 & 4.4 \\
SynthSR & 21.81 & 0.704 & 0.094 & 0.188 & 0.605 & 48.4 \\
\midrule
\multicolumn{7}{l}{\textit{Through-plane $3\times$}} \\
GSMR3D (ours) & 33.37 & 0.981 & \textbf{0.020} & \textbf{0.060} & 0.806 & \textbf{4.9} \\
Diffusion prior + DC & 32.95 & 0.979 & 0.034 & 0.079 & 0.805 & 10.3 \\
SAINT (specialist) & \textbf{34.30} & \textbf{0.983} & 0.031 & 0.077 & \textbf{0.834} & 10.5 \\
TV & 31.37 & 0.971 & 0.052 & 0.096 & 0.757 & 18.4 \\
Cubic & 29.57 & 0.956 & 0.102 & 0.125 & 0.701 & 21.3 \\
ResCNN model & 33.49 & 0.982 & 0.033 & 0.080 & 0.819 & 11.4 \\
SynthSR & 21.81 & 0.682 & 0.109 & 0.200 & 0.599 & 54.7 \\
\midrule
\multicolumn{7}{l}{\textit{Through-plane $4\times$}} \\
GSMR3D (ours) & 31.15 & 0.968 & \textbf{0.039} & \textbf{0.083} & 0.749 & \textbf{9.1} \\
Diffusion prior + DC & 30.70 & 0.965 & 0.053 & 0.099 & 0.745 & 13.5 \\
SAINT (specialist) & \textbf{31.98} & \textbf{0.972} & 0.052 & 0.098 & \textbf{0.776} & 15.7 \\
TV & 29.36 & 0.954 & 0.087 & 0.123 & 0.689 & 28.6 \\
Cubic & 27.93 & 0.935 & 0.140 & 0.148 & 0.642 & 29.6 \\
ResCNN model & 31.25 & 0.969 & 0.054 & 0.101 & 0.760 & 15.4 \\
SynthSR & 21.57 & 0.670 & 0.123 & 0.208 & 0.583 & 59.2 \\
\bottomrule
\end{tabular}
\end{adjustbox}
\end{table*}
\begin{table*}[!tbp]
\centering
\caption{\textbf{Image quality analysis on T1 volumes (brain-metastasis cohort, $n{=}100$).} Same protocol as Table~\ref{tab:t1}}
\label{tab:bm-t1}
\begin{adjustbox}{max width=0.8\linewidth}
\footnotesize
\renewcommand{\arraystretch}{0.9}
\setlength{\tabcolsep}{5pt}
\begin{tabular}{lcccccc}
\toprule
Method & PSNR3D$\uparrow$ & SSIM3D$\uparrow$ & LPIPS$\downarrow$ & DISTS$\downarrow$ & HaarPSI$\uparrow$ & FID$\downarrow$ \\
\midrule
\multicolumn{7}{l}{\textit{Isotropic $2\times$}} \\
GSMR3D (ours) & \textbf{33.84} & \textbf{0.982} & \textbf{0.026} & \textbf{0.067} & \textbf{0.804} & \textbf{6.3} \\
Diffusion prior + DC & 32.47 & 0.970 & 0.042 & 0.103 & 0.778 & 10.8 \\
ArSSR (specialist) & 32.98 & 0.978 & 0.035 & 0.086 & 0.740 & 11.5 \\
TV & 31.10 & 0.969 & 0.054 & 0.099 & 0.726 & 14.7 \\
Cubic & 27.91 & 0.935 & 0.131 & 0.136 & 0.652 & 28.4 \\
ResCNN model & 32.50 & 0.976 & 0.040 & 0.092 & 0.772 & 12.5 \\
SynthSR & 24.50 & 0.752 & 0.105 & 0.181 & 0.628 & 49.6 \\
\midrule
\multicolumn{7}{l}{\textit{In-plane $2\times$}} \\
GSMR3D (ours) & 34.81 & \textbf{0.986} & \textbf{0.015} & \textbf{0.051} & \textbf{0.834} & \textbf{2.6} \\
Diffusion prior + DC & 33.95 & 0.981 & 0.025 & 0.067 & 0.819 & 5.8 \\
LTE (specialist) & \textbf{34.88} & 0.985 & 0.019 & 0.057 & 0.804 & 5.0 \\
TV & 32.48 & 0.978 & 0.034 & 0.075 & 0.780 & 8.0 \\
Cubic & 29.42 & 0.955 & 0.091 & 0.114 & 0.706 & 19.1 \\
ResCNN model & 34.21 & 0.983 & 0.022 & 0.065 & 0.819 & 6.4 \\
SynthSR & 24.76 & 0.761 & 0.090 & 0.171 & 0.638 & 45.8 \\
\midrule
\multicolumn{7}{l}{\textit{Through-plane $3\times$}} \\
GSMR3D (ours) & 32.89 & 0.978 & \textbf{0.024} & \textbf{0.070} & 0.787 & \textbf{6.7} \\
Diffusion prior + DC & 32.97 & 0.973 & 0.036 & 0.103 & 0.798 & 12.6 \\
SAINT (specialist) & \textbf{34.17} & \textbf{0.984} & 0.030 & 0.081 & \textbf{0.832} & 11.7 \\
TV & 31.16 & 0.972 & 0.056 & 0.103 & 0.750 & 23.1 \\
Cubic & 29.38 & 0.957 & 0.106 & 0.132 & 0.699 & 26.9 \\
ResCNN model & 32.55 & 0.979 & 0.037 & 0.089 & 0.788 & 12.8 \\
SynthSR & 24.50 & 0.747 & 0.105 & 0.183 & 0.632 & 50.2 \\
\midrule
\multicolumn{7}{l}{\textit{Through-plane $4\times$}} \\
GSMR3D (ours) & 30.53 & 0.966 & \textbf{0.041} & \textbf{0.092} & 0.731 & \textbf{14.4} \\
Diffusion prior + DC & 30.71 & 0.961 & 0.056 & 0.124 & 0.742 & 16.6 \\
SAINT (specialist) & \textbf{31.83} & \textbf{0.974} & 0.048 & 0.100 & \textbf{0.778} & 16.9 \\
TV & 29.13 & 0.955 & 0.092 & 0.131 & 0.682 & 34.4 \\
Cubic & 27.73 & 0.936 & 0.145 & 0.155 & 0.640 & 35.8 \\
ResCNN model & 30.42 & 0.966 & 0.054 & 0.106 & 0.730 & 15.9 \\
SynthSR & 24.14 & 0.734 & 0.119 & 0.192 & 0.612 & 55.2 \\
\bottomrule
\end{tabular}
\end{adjustbox}
\end{table*}

\subsection{Baseline implementation and training}
\label{app:baseline_training}

All baselines trained in this study use the same data partitions as GSMR3D and the same Gaussian anti-aliasing and subsampling operator for generating LR inputs. Separate models are trained for T1 and CT1. All learned baselines, are trained for 600 epochs, with the checkpoint achieving the lowest validation loss selected for evaluation. For T1, trainable baselines follow the same sequential training protocol as GSMR3D, with initial training on NAKO followed by fine-tuning on CENTRIC and CORE. ArSSR \citep{wu2023arssr} is trained for arbitrary-scale isotropic SR and evaluated at isotropic $2\times$. LTE \citep{lee2022lte} is trained as a 2D arbitrary-scale model on axial slices and evaluated for in-plane $2\times$ SR. SAINT \citep{peng2020saint} uses its two-stage sagittal/coronal reconstruction and fusion framework and is evaluated for through-plane $3\times$ and $4\times$ SR. ResCNN is based on \citet{du2020super} and is trained with the same independently sampled per-axis $1$--$4\times$ scale distribution as GSMR3D, providing a conventional 3D CNN baseline that uses a single model per contrast across isotropic and anisotropic geometries.

Our diffusion baseline, WaveNCSN, is based on the NCSN++ architecture \citep{song2021scorebased}. Rather than operating directly on MRI volumes, it operates on a level-1 Haar wavelet decomposition, which transforms each single-channel volume into eight channels at half the spatial resolution along each axis. This increases the effective receptive field while reducing memory requirements. For reconstruction, the unconditional diffusion prior is combined with DDNM+ \citep{wang2023zeroshot}. Successive score evaluations are performed along the three anatomical orientations in round-robin order. We use a time-travel length of 10 steps, an interval of 10 steps, and five repeats, resulting in 195 score evaluations per reconstruction. These settings were selected based on preliminary experiments.

Total-variation reconstruction and cubic interpolation require no training. SynthSR \citep{iglesias2023synthsr} is evaluated using its publicly released pretrained weights without fine-tuning.
\section{Additional results}
\label{app:results}

\begin{table*}[!tbp]
\centering
\caption{\textbf{Image quality analysis on contrast-enhanced T1 volumes (glioblastoma cohort, $n{=}100$).} One GSMR3D checkpoint against the geometry-specific specialist of each geometry, a wavelet-domain diffusion prior sampled with data consistency, and classical baselines. Bold marks the best among the methods of Table~\ref{tab:ct1}.}
\label{tab:ct1-full}
\begin{adjustbox}{max width=0.8\linewidth}
\footnotesize
\renewcommand{\arraystretch}{0.9}
\setlength{\tabcolsep}{5pt}
\begin{tabular}{lcccccc}
\toprule
Method & PSNR3D$\uparrow$ & SSIM3D$\uparrow$ & LPIPS$\downarrow$ & DISTS$\downarrow$ & HaarPSI$\uparrow$ & FID$\downarrow$ \\
\midrule
\multicolumn{7}{l}{\textit{Isotropic $2\times$}} \\
GSMR3D (ours) & \textbf{32.50} & \textbf{0.969} & \textbf{0.029} & \textbf{0.074} & \textbf{0.793} & \textbf{5.6} \\
Diffusion prior + DC & 31.84 & 0.963 & 0.045 & 0.085 & 0.781 & 6.3 \\
ArSSR (specialist) & 32.24 & 0.964 & 0.043 & 0.090 & 0.706 & 9.4 \\
TV & 30.22 & 0.952 & 0.065 & 0.104 & 0.718 & 11.6 \\
Cubic & 26.98 & 0.909 & 0.147 & 0.140 & 0.647 & 20.8 \\
ResCNN model & 32.47 & 0.967 & 0.045 & 0.087 & 0.790 & 8.2 \\
 SynthSR & 20.81 & 0.718 & 0.143 & 0.205 & 0.585 & 50.6  \\
\midrule
\multicolumn{7}{l}{\textit{In-plane $2\times$}} \\
GSMR3D (ours) & 32.69 & 0.974 & \textbf{0.020} & \textbf{0.063} & 0.802 & \textbf{3.7} \\
Diffusion prior + DC & 32.89 & 0.972 & 0.031 & 0.070 & 0.808 & 4.4 \\
LTE (specialist) & \textbf{33.72} & \textbf{0.975} & 0.027 & 0.067 & \textbf{0.825} & 4.1 \\
TV & 31.10 & 0.964 & 0.044 & 0.086 & 0.736 & 6.8 \\
Cubic & 28.22 & 0.934 & 0.106 & 0.123 & 0.689 & 15.2 \\
ResCNN model & 33.73 & 0.975 & 0.029 & 0.069 & 0.822 & 5.3 \\
SynthSR & 25.82 & 0.904 & 0.088 & 0.152 & 0.586 & 28.5 \\

\midrule
\multicolumn{7}{l}{\textit{Through-plane $3\times$}} \\
GSMR3D (ours) & 32.95 & 0.973 & \textbf{0.025} & \textbf{0.067} & 0.817 & \textbf{7.2} \\
Diffusion prior + DC & 33.14 & 0.972 & 0.038 & 0.080 & 0.823 & 10.3 \\
SAINT (specialist) & \textbf{34.36} & \textbf{0.976} & 0.036 & 0.081 & \textbf{0.844} & 11.2 \\
TV & 31.22 & 0.963 & 0.061 & 0.096 & 0.776 & 20.8 \\
Cubic & 29.12 & 0.944 & 0.118 & 0.128 & 0.720 & 23.3 \\
ResCNN model & 33.50 & 0.975 & 0.038 & 0.084 & 0.828 & 12.7 \\
SynthSR & 20.59 & 0.718 & 0.143 & 0.209 & 0.586 & 51.9  \\
\midrule
\multicolumn{7}{l}{\textit{Through-plane $4\times$}} \\
GSMR3D (ours) & 30.86 & 0.959 & \textbf{0.045} & \textbf{0.088} & 0.764 & \textbf{10.8} \\
Diffusion prior + DC & 30.69 & 0.955 & 0.060 & 0.100 & 0.764 & 12.8 \\
SAINT (specialist) & \textbf{31.92} & \textbf{0.962} & 0.059 & 0.103 & \textbf{0.789} & 16.2 \\
TV & 28.89 & 0.941 & 0.104 & 0.125 & 0.706 & 32.1 \\
Cubic & 27.33 & 0.919 & 0.162 & 0.155 & 0.662 & 31.4 \\
ResCNN model & 31.15 & 0.960 & 0.061 & 0.106 & 0.774 & 17.0 \\
SynthSR & 20.42 & 0.709 & 0.157 & 0.215 & 0.574 & 56.4  \\
\bottomrule
\end{tabular}
\end{adjustbox}
\end{table*}

\begin{table*}[!tbp]
\centering
\caption{\textbf{Image quality analysis on T1 volumes (NAKO $+$ glioblastoma cohort, $n{=}200$).} Same protocol as Table~\ref{tab:ct1}.}
\label{tab:t1}
\begin{adjustbox}{max width=0.8\linewidth}
\footnotesize
\setlength{\tabcolsep}{5pt}
\begin{tabular}{lcccccc}
\toprule
Method & PSNR3D$\uparrow$ & SSIM3D$\uparrow$ & LPIPS$\downarrow$ & DISTS$\downarrow$ & HaarPSI$\uparrow$ & FID$\downarrow$ \\
\midrule
\multicolumn{7}{l}{\textit{Isotropic $2\times$}} \\
GSMR3D (ours) & \textbf{34.71} & \textbf{0.975} & \textbf{0.030} & \textbf{0.072} & \textbf{0.826} & \textbf{5.3} \\
Diffusion prior + DC & 33.49 & 0.964 & 0.047 & 0.102 & 0.801 & 7.2 \\
ArSSR (specialist) & 33.61 & 0.969 & 0.042 & 0.087 & 0.732 & 8.8 \\
TV & 32.05 & 0.960 & 0.064 & 0.101 & 0.750 & 10.8 \\
Cubic & 28.65 & 0.927 & 0.139 & 0.137 & 0.686 & 21.8 \\
ResCNN model & 32.96 & 0.967 & 0.048 & 0.096 & 0.788 & 10.6 \\
 SynthSR & 25.17 & 0.797 & 0.129 & 0.190 & 0.631 & 53.9 \\
\midrule
\multicolumn{7}{l}{\textit{In-plane $2\times$}} \\
GSMR3D (ours) & 34.99 & \textbf{0.979} & \textbf{0.018} & \textbf{0.057} & \textbf{0.840} & \textbf{3.0} \\
Diffusion prior + DC & 34.41 & 0.974 & 0.032 & 0.074 & 0.827 & 5.1 \\
LTE (specialist) & \textbf{35.43} & 0.978 & 0.027 & 0.065 & 0.796 & 4.4 \\
TV & 32.96 & 0.970 & 0.043 & 0.084 & 0.772 & 6.9 \\
Cubic & 29.92 & 0.946 & 0.100 & 0.121 & 0.729 & 16.7 \\
ResCNN model & 34.31 & 0.975 & 0.030 & 0.075 & 0.822 & 6.3 \\
SynthSR & 24.40 & 0.782 & 0.116 & 0.179 & 0.646 & 41.6 \\
\midrule
\multicolumn{7}{l}{\textit{Through-plane $3\times$}} \\
GSMR3D (ours) & 34.11 & 0.976 & \textbf{0.030} & 0.074 & 0.826 & \textbf{8.7} \\
Diffusion prior + DC & 35.26 & 0.973 & 0.034 & 0.090 & 0.845 & \textbf{8.7} \\
SAINT (specialist) & \textbf{35.45} & \textbf{0.980} & \textbf{0.030} & \textbf{0.072} & \textbf{0.868} & 9.5 \\
TV & 33.12 & 0.970 & 0.054 & 0.087 & 0.803 & 18.1 \\
Cubic & 30.80 & 0.954 & 0.109 & 0.120 & 0.749 & 22.5 \\
ResCNN model & 33.63 & 0.975 & 0.041 & 0.087 & 0.818 & 13.3 \\
SynthSR & 24.99 & 0.791 & 0.116 & 0.185 & 0.630 & 53.6 \\
\midrule
\multicolumn{7}{l}{\textit{Through-plane $4\times$}} \\
GSMR3D (ours) & 32.15 & 0.964 & \textbf{0.044} & \textbf{0.092} & 0.773 & 13.3 \\
Diffusion prior + DC & 32.68 & 0.959 & 0.055 & 0.112 & 0.787 & \textbf{11.4} \\
SAINT (specialist) & \textbf{33.15} & \textbf{0.970} & 0.051 & 0.094 & \textbf{0.816} & 14.1 \\
TV & 30.61 & 0.952 & 0.093 & 0.116 & 0.733 & 29.0 \\
Cubic & 28.89 & 0.932 & 0.150 & 0.146 & 0.691 & 29.2 \\
ResCNN model & 31.76 & 0.961 & 0.060 & 0.104 & 0.768 & 16.0 \\
SynthSR & 24.70 & 0.786 & 0.136 & 0.201 & 0.623 & 56.5 \\
\bottomrule
\end{tabular}
\end{adjustbox}
\end{table*}

\subsection{Image quality metrics}
\label{app:metrics}

We evaluate reconstruction quality using PSNR and SSIM, computed in 3D over the full volume. We additionally report HaarPSI \citep{reisenhofer:haarpsi} and DISTS \citep{ding:dists}, which have been shown to correlate well with diagnostic image quality in MRI \citep{kastryulin:iqaformri}. Since these metrics operate on 2D images, we compute them for each axial slice and average the scores across the volume. Finally, we report LPIPS \citep{zhang:lpips} and FID \citep{heusel:ganstrained}. LPIPS is computed for each axial slice and averaged over the volume, while FID is computed over all reconstructed and corresponding ground-truth axial slices. Tables~\ref{tab:ct1-full} and \ref{tab:t1} report the complete results for CT1 and T1, respectively.

\subsection{Runtime}
\label{app:runtime}

Table~\ref{tab:runtime} reports the median end-to-end reconstruction time per volume. GPU-based methods are evaluated on a single NVIDIA A100 GPU. The reported diffusion runtime corresponds to a single posterior sample and increases linearly with the number of samples. SynthSR was evaluated on CPU because its released TensorFlow implementation was incompatible with our available GPU environment, and its runtime is therefore not directly comparable to the GPU measurements.

\begin{table*}[!tbp]
\centering
\caption{\textbf{Inference time per volume} (median time on a single NVIDIA A100 GPU). GSMR3D is end-to-end, including encoding, all HR queries through the fused gather kernel, and the measurement projection. Specialists exist only for their geometry. SynthSR was run on CPU because its released package does not support our GPU architecture. 
}
\label{tab:runtime}
\footnotesize
\renewcommand{\arraystretch}{0.9}
\setlength{\tabcolsep}{7pt}
\begin{adjustbox}{max width=\linewidth}
\begin{tabular}{lcccc}
\toprule
Method & Isotropic $2\times$ & In-plane $2\times$ & Through-plane $3\times$ & Through-plane $4\times$ \\
\midrule
GSMR3D (ours, end-to-end) & 6.4 & 5.8 & 5.9 & 6.1 \\
Specialist (ArSSR / LTE / SAINT) & 8.2 & 19.7 & 5.9 & 5.3 \\
Diffusion prior + DC & 140.3 & 149.0 & 137.6 & 136.4 \\
TV & 31.1 & 31.3 & 31.4 & 31.1 \\
Interpolation & 8.2 & 7.4 & 8.5 & 8.0 \\
SynthSR (CPU) & 801.3 & -- & 808.3 & 805.8 \\
\midrule
ResCNN (residual CNN) & 6.8 & 8.6 & 7.0 & 6.8 \\
\bottomrule
\end{tabular}
\end{adjustbox}
\end{table*}

\subsection{Downstream tumour segmentation}
\label{app:seg}

Table~\ref{tab:seg-full} extends the results in Table~\ref{tab:seg} with the non-enhancing tumour label, Dice against expert annotations, and tumour-volume errors. Agreement with expert annotations differs less between reconstruction methods because segmentation of the original HR volumes already achieves Dice scores of 0.802, 0.911, and 0.927 for contrast-enhancing (CE), non-enhancing (NE), and whole tumour (WT), respectively. At isotropic $2\times$, ArSSR achieves slightly higher CE Dice against the expert annotation than GSMR3D (0.800 vs.\ 0.796), while GSMR3D shows higher agreement with the segmentation obtained from the HR volume (0.961 vs.\ 0.955) and lower tumour-volume error (0.5\% vs.\ 0.9\%). Across the remaining geometries, GSMR3D achieves the highest agreement across all three segmentation measures.

Paired Wilcoxon tests on CE DICE agreement for the cases containing contrast-enhancing tumour further compare GSMR3D with ResCNN, the geometry-specific specialist, and the diffusion prior. GSMR3D achieves higher per-case agreement across all four geometries, with $p\leq0.04$ for all 12 comparisons and $p<10^{-3}$ for 10 of them. Depending on the geometry and comparator, GSMR3D achieves higher agreement in 64--95\% of individual cases.

\begin{table*}[!tbp]
\centering
\caption{\textbf{Full downstream tumour-segmentation results} on contrast-enhanced T1-weighted images from the held-out multicentre brain-tumour cohort using HD-GLIO. \emph{Agreement} is the Dice score between segmentations obtained from the SR and HR volumes. \emph{vs.\ GT} is the Dice score against expert annotations. For reference, HD-GLIO applied to the HR images achieves 0.802, 0.911, and 0.927 for CE, NE, and WT, respectively. \emph{Vol.\ err.} is the median absolute relative error in tumour volume with respect to the HR-derived segmentation. The best result in each column and geometry is shown in bold.}
\label{tab:seg-full}
\footnotesize
\renewcommand{\arraystretch}{0.9}
\setlength{\tabcolsep}{3.5pt}
\begin{tabular}{llcccccccc}
\toprule
 & & \multicolumn{3}{c}{Agreement Dice$\uparrow$} & \multicolumn{3}{c}{Dice vs.\ GT$\uparrow$} & \multicolumn{2}{c}{Vol.\ err.\ (\%)$\downarrow$} \\
\cmidrule(lr){3-5}\cmidrule(lr){6-8}\cmidrule(lr){9-10}
Geometry & Method & CE & NE & WT & CE & NE & WT & CE & WT \\
\midrule
Iso $2\times$ & GSMR3D (ours) & \textbf{0.961} & \textbf{0.993} & \textbf{0.994} & 0.796 & \textbf{0.911} & \textbf{0.927} & \textbf{0.5} & \textbf{0.3} \\
 & Diffusion prior + DC & 0.946 & 0.991 & 0.992 & 0.791 & 0.910 & 0.926 & 0.9 & 0.5 \\
 & ArSSR (specialist) & 0.955 & 0.990 & 0.992 & \textbf{0.800} & 0.911 & 0.927 & 0.9 & 0.4 \\
 & TV & 0.941 & 0.987 & 0.989 & 0.789 & 0.909 & 0.926 & 1.7 & 0.8 \\
 & Cubic & 0.859 & 0.962 & 0.968 & 0.764 & 0.900 & 0.920 & 13.7 & 2.0 \\
 & ResCNN & 0.937& 0.990 & 0.992 & 0.785 & 0.911 & 0.927 & 0.6 & 0.5 \\
 & SynthSR & 0.113 & 0.750 & 0.802 & 0.107 & 0.732 & 0.783 & 83.6 & 23.0 \\
\midrule
In-plane $2\times$ & GSMR3D (ours) & \textbf{0.967} & \textbf{0.994} & \textbf{0.995} & 0.797 & \textbf{0.911} & \textbf{0.927} & 0.7 & \textbf{0.2} \\
 & Diffusion prior + DC & 0.956 & 0.992 & 0.993 & 0.793 & 0.911 & 0.927 & 0.8 & 0.4 \\
 & LTE (specialist) & 0.966 & 0.993 & 0.994 & \textbf{0.798} & 0.911 & 0.927 & \textbf{0.5} & 0.4 \\
 & TV & 0.951 & 0.990 & 0.991 & 0.792 & 0.910 & 0.926 & 1.5 & 0.6 \\
 & Cubic & 0.901 & 0.969 & 0.974 & 0.779 & 0.903 & 0.921 & 8.5 & 1.8 \\
 & ResCNN & 0.962 & 0.992 & 0.994 & 0.795 & 0.911 & 0.927 & 0.6 & 0.4\\
 & SynthSR & 0.105 & 0.750 & 0.804 & 0.101 & 0.732 & 0.785 & 82.4 & 22.0 \\  
\midrule
Through $3\times$ & GSMR3D (ours) & \textbf{0.947} & \textbf{0.991} & \textbf{0.992} & \textbf{0.791} & \textbf{0.911} & \textbf{0.927} & 0.9 & \textbf{0.3} \\
 & Diffusion prior + DC & 0.928 & 0.988 & 0.990 & 0.785 & 0.909 & 0.926 & 1.4 & \textbf{0.3} \\
 & SAINT (specialist) & 0.933 & 0.989 & 0.991 & 0.788 & 0.910 & 0.926 & 1.0 & \textbf{0.3} \\
 & TV & 0.914 & 0.986 & 0.988 & 0.782 & 0.909 & 0.926 & 2.9 & 0.5 \\
 & Cubic & 0.870 & 0.972 & 0.977 & 0.771 & 0.906 & 0.925 & 11.3 & 1.8 \\
 & ResCNN & 0.934 & 0.988 & 0.990 & 0.787 & 0.910 & 0.926 & \textbf{0.8} & 0.5 \\
 & SynthSR & 0.109 & 0.755 & 0.806 & 0.105 & 0.737 & 0.787 & 77.4 & 21.6 \\
\midrule
Through $4\times$ & GSMR3D (ours) & \textbf{0.905} & \textbf{0.983} & \textbf{0.986} & \textbf{0.781} & \textbf{0.908} & \textbf{0.926} & 2.3 & 0.6 \\
 & Diffusion prior + DC & 0.886 & 0.979 & 0.983 & 0.773 & 0.907 & 0.925 & 2.9 & 0.7 \\
 & SAINT (specialist) & 0.900 & 0.981 & 0.985 & 0.779 & 0.908 & 0.925 & 2.5 & \textbf{0.5} \\
 & TV & 0.875 & 0.976 & 0.980 & 0.771 & 0.906 & 0.925 & 7.1 & 0.9 \\
 & Cubic & 0.824 & 0.958 & 0.967 & 0.753 & 0.901 & 0.922 & 16.3 & 2.9 \\
 & ResCNN & 0.893 & 0.980 & 0.983 & 0.776 & 0.907 & 0.925 & \textbf{2.1} & 0.6 \\
& SynthSR & 0.110 & 0.753 & 0.803 & 0.105 & 0.735 & 0.785 & 79.5 & 22.6 \\  
\bottomrule
\end{tabular}
\end{table*}

\subsection{Downstream brain parcellation}
\label{app:parc}

Table~\ref{tab:parc-full} reports agreement separately for each brain structure in the healthy T1 cohort. At isotropic and in-plane $2\times$, GSMR3D achieves the highest Dice or remains within 0.001 of the highest value for every structure. At through-plane $3\times$ and $4\times$, the geometry-specific specialist and ResCNN perform better for the cerebral cortex, cerebral white matter, CSF, and lateral ventricles, while GSMR3D leads or remains comparable for several subcortical structures, including the putamen and hippocampus, as well as the brainstem. GSMR3D achieves higher mean agreement than the diffusion prior across all four geometries.

\begin{table*}[!tbp]
\centering
\caption{\textbf{Per-structure brain parcellation agreement} (T1, NAKO, $n{=}100$): Dice between the parcellation of the super-resolved and of the true HR volume. All = mean over 20 structures. Best per column within each geometry in bold.}
\label{tab:parc-full}
\begin{adjustbox}{max width=\linewidth}
\footnotesize
\renewcommand{\arraystretch}{0.9}
\setlength{\tabcolsep}{3pt}
\begin{tabular}{llcccccccccc}
\toprule
Geometry & Method & All & Cortex & WM & CSF & Lat.\ vent. & Hippoc. & Thal. & Putamen & Cereb.\ ctx & Brainstem \\
\midrule
Iso $2\times$ & GSMR3D (ours) & \textbf{0.940} & \textbf{0.953} & \textbf{0.973} & \textbf{0.921} & \textbf{0.979} & \textbf{0.950} & \textbf{0.965} & \textbf{0.951} & \textbf{0.956} & \textbf{0.973} \\
 & Diffusion prior + DC & 0.932 & 0.939 & 0.967 & 0.902 & 0.975 & 0.943 & 0.962 & 0.946 & 0.946 & 0.971 \\
 & ArSSR (specialist) & 0.931 & 0.947 & 0.969 & 0.910 & 0.975 & 0.943 & 0.961 & 0.944 & 0.946 & 0.969 \\
 & TV & 0.919 & 0.910 & 0.956 & 0.869 & 0.968 & 0.930 & 0.956 & 0.943 & 0.930 & 0.964 \\
 & Cubic & 0.915 & 0.903 & 0.954 & 0.857 & 0.957 & 0.925 & 0.957 & 0.942 & 0.924 & 0.967 \\
 & ResCNN & 0.930 & 0.944 & 0.968 & 0.906 & 0.975 & 0.941 & 0.962 & 0.946 & 0.943 & 0.970 \\
  & SynthSR & 0.898 & 0.880 & 0.936 & 0.831 & 0.953 & 0.926 & 0.956 & 0.935 & 0.905 & 0.958 \\
\midrule
In-plane $2\times$ & GSMR3D (ours) & \textbf{0.956} & \textbf{0.962} & \textbf{0.979} & \textbf{0.936} & \textbf{0.984} & \textbf{0.968} & \textbf{0.974} & \textbf{0.964} & \textbf{0.969} & \textbf{0.979} \\
 & Diffusion prior + DC & 0.951 & 0.955 & 0.975 & 0.927 & 0.982 & 0.963 & 0.971 & 0.960 & 0.964 & 0.977 \\
 & LTE (specialist) & 0.954 & 0.961 & 0.978 & 0.935 & \textbf{0.984} & 0.965 & 0.972 & 0.961 & 0.966 & \textbf{0.979} \\
 & TV & 0.941 & 0.938 & 0.968 & 0.904 & 0.976 & 0.957 & 0.965 & 0.957 & 0.957 & 0.971 \\
 & Cubic & 0.931 & 0.922 & 0.962 & 0.882 & 0.968 & 0.946 & 0.963 & 0.953 & 0.946 & 0.969 \\
 & ResCNN & 0.950 & 0.958 & 0.976 & 0.930 & 0.982 & 0.963 & 0.971 & 0.959 & 0.962 & 0.977 \\
 & SynthSR & 0.904 & 0.892 & 0.942 & 0.843 & 0.958 & 0.932 & 0.957 & 0.939 & 0.913 & 0.959 \\
\midrule
Through $3\times$ & GSMR3D (ours) & 0.940 & 0.943 & 0.968 & 0.908 & 0.965 & \textbf{0.945} & 0.967 & \textbf{0.961} & \textbf{0.952} & \textbf{0.982} \\
 & Diffusion prior + DC & 0.938 & 0.941 & 0.967 & 0.909 & 0.974 & 0.938 & 0.969 & 0.957 & 0.946 & 0.981 \\
 & SAINT (specialist) & \textbf{0.943} & \textbf{0.954} & \textbf{0.973} & \textbf{0.924} & \textbf{0.979} & \textbf{0.945} & \textbf{0.970} & 0.956 & \textbf{0.952} & 0.981 \\
 & TV & 0.922 & 0.914 & 0.953 & 0.873 & 0.960 & 0.924 & 0.963 & 0.951 & 0.928 & 0.977 \\
 & Cubic & 0.919 & 0.907 & 0.951 & 0.861 & 0.955 & 0.921 & 0.960 & 0.948 & 0.924 & 0.975 \\
 & ResCNN & 0.942 & 0.953 & 0.973 & 0.922 & 0.979 & 0.945 & 0.970 & 0.956 & 0.951 & 0.981 \\
  & SynthSR & 0.897 & 0.884 & 0.937 & 0.831 & 0.952 & 0.921 & 0.956 & 0.934 & 0.900 & 0.963 \\
\midrule
Through $4\times$ & GSMR3D (ours) & 0.923 & 0.921 & 0.956 & 0.878 & 0.956 & 0.924 & 0.959 & \textbf{0.951} & \textbf{0.938} & \textbf{0.977} \\
 & Diffusion prior + DC & 0.921 & 0.915 & 0.953 & 0.877 & 0.966 & 0.922 & 0.963 & 0.947 & 0.932 & 0.976 \\
 & SAINT (specialist) & \textbf{0.928} & \textbf{0.936} & \textbf{0.964} & \textbf{0.898} & \textbf{0.972} & \textbf{0.930} & \textbf{0.964} & 0.946 & \textbf{0.938} & 0.976 \\
 & TV & 0.901 & 0.877 & 0.934 & 0.828 & 0.945 & 0.900 & 0.953 & 0.939 & 0.908 & 0.970 \\
 & Cubic & 0.901 & 0.875 & 0.933 & 0.820 & 0.941 & 0.902 & 0.953 & 0.939 & 0.909 & 0.970 \\
  & ResCNN & 0.926 & 0.936 & 0.964 & 0.896 & 0.970 & 0.927 & 0.963 & 0.945 & 0.936 & \textbf{0.977} \\
  & SynthSR & 0.889 & 0.868 & 0.928 & 0.814 & 0.945 & 0.913 & 0.954 & 0.932 & 0.892 & 0.962 \\
\bottomrule
\end{tabular}
\end{adjustbox}
\end{table*}

\subsection{Downstream brain-metastasis segmentation}
\label{app:bm-seg}

For the external pathology evaluation, we select 100 brain-metastasis cases from the MR-RATE dataset \citep{mrrate} with T1-weighted, contrast-enhanced T1-weighted, and FLAIR sequences. The cohort contains 670 enhancing metastases and is used exclusively for evaluation, without additional training or fine-tuning of GSMR3D. We apply HD-BM \citep{hd-bm} to the reconstructed and original HR volumes and measure the agreement between their enhancing-metastasis (CE) and FLAIR-abnormality (FL) segmentations. To assess preservation of individual lesions, we additionally perform lesion-level analysis against the HR-derived segmentation, defining lesions as 26-connected components containing at least five voxels. We report lesion sensitivity, the total number of missed lesions, false positives per case, and the median absolute relative error in enhancing-lesion volume.

\begin{table*}[!tbp]
\centering
\caption{\textbf{Downstream brain-metastasis evaluation} on the held-out cohort ($n{=}100$, 670 enhancing metastases) using HD-BM. Agreement is measured between SR- and HR-derived segmentations for enhancing metastases (CE) and FLAIR abnormalities (FL). Lesion sensitivity, missed lesions, false positives per case, and enhancing-volume error are evaluated relative to the HR-derived segmentation. Best results for each geometry are shown in bold.}
\label{tab:bm-seg}
\footnotesize
\renewcommand{\arraystretch}{0.9}
\setlength{\tabcolsep}{3.5pt}
\begin{tabular}{llccccccc}
\toprule
 & & \multicolumn{2}{c}{Agreement Dice$\uparrow$} & \multicolumn{3}{c}{Lesion detection (vs.\ HR)} & Vol.\ err.\ (\%)$\downarrow$ \\
\cmidrule(lr){3-4}\cmidrule(lr){5-7}\cmidrule(lr){8-8}
Geometry & Method & CE & FL & Sens.$\uparrow$ & Missed$\downarrow$ & FP/case$\downarrow$ & CE \\
\midrule
Iso $2\times$ & GSMR3D (ours) & \textbf{0.939} & 0.978 & \textbf{0.994} & \textbf{4} & 0.03 & \textbf{2.2} \\
 & Diffusion prior + DC & 0.927 & \textbf{0.981} & 0.982 & 12 & \textbf{0.02} & 3.6 \\
 & ArSSR (specialist) & 0.927 & 0.966 & 0.984 & 11 & 0.04 & 2.6 \\
 & TV & 0.918 & 0.960 & 0.976 & 16 & 0.03 & 4.1 \\
 & Cubic & 0.852 & 0.911 & 0.964 & 24 & 0.24 & 14.7 \\
 & ResCNN & 0.920 & 0.970 & 0.984 & 11 & \textbf{0.02} & 2.4 \\
 & SynthSR & 0.233 & 0.590 & 0.376 & 418 & 2.76 & 65.1 \\
\midrule
In-plane $2\times$ & GSMR3D (ours) & 0.949 & 0.982 & \textbf{0.996} & \textbf{3} & 0.01 & 1.4 \\
 & Diffusion prior + DC & 0.950 & \textbf{0.985} & 0.994 & 4 & \textbf{0.00} & 1.8 \\
 & LTE (specialist) & \textbf{0.953} & 0.983 & 0.993 & 5 & \textbf{0.00} & \textbf{1.3} \\
 & TV & 0.941 & 0.982 & 0.991 & 6 & 0.02 & 2.1 \\
 & Cubic & 0.874 & 0.924 & 0.969 & 21 & 0.13 & 9.7 \\
 & ResCNN & 0.947 & 0.977 & \textbf{0.996} & \textbf{3} & \textbf{0.00} & 1.8 \\
 & SynthSR & 0.218 & 0.607 & 0.345 & 439 & 3.16 & 65.2 \\
\midrule
Through $3\times$ & GSMR3D (ours) & \textbf{0.911} & \textbf{0.962} & \textbf{0.970} & \textbf{20} & \textbf{0.07} & \textbf{2.7} \\
 & Diffusion prior + DC & 0.880 & 0.955 & 0.943 & 38 & 0.08 & 3.2 \\
 & SAINT (specialist) & 0.903 & \textbf{0.962} & 0.969 & 21 & 0.08 & 3.2 \\
 & TV & 0.872 & 0.933 & 0.945 & 37 & 0.10 & 4.8 \\
 & Cubic & 0.837 & 0.901 & 0.934 & 44 & 0.18 & 9.6 \\
 & ResCNN & 0.896 & 0.957 & 0.954 & 31 & \textbf{0.07} & 3.8 \\
 & SynthSR & 0.204 & 0.602 & 0.336 & 445 & 2.97 & 69.9 \\
\midrule
Through $4\times$ & GSMR3D (ours) & \textbf{0.838} & \textbf{0.931} & \textbf{0.915} & \textbf{57} & 0.10 & \textbf{4.1} \\
 & Diffusion prior + DC & 0.819 & 0.922 & 0.867 & 89 & \textbf{0.05} & 6.5 \\
 & SAINT (specialist) & 0.835 & \textbf{0.931} & 0.907 & 62 & 0.11 & 5.9 \\
 & TV & 0.803 & 0.894 & 0.897 & 69 & 0.09 & 7.0 \\
 & Cubic & 0.768 & 0.876 & 0.861 & 93 & 0.27 & 13.7 \\
 & ResCNN & 0.821 & 0.924 & 0.873 & 85 & 0.09 & 6.3 \\
 & SynthSR & 0.195 & 0.588 & 0.309 & 463 & 2.81 & 70.6 \\
\bottomrule
\end{tabular}
\end{table*}

As shown in Table~\ref{tab:bm-seg}, GSMR3D achieves the highest CE segmentation agreement at isotropic $2\times$ and both through-plane geometries, while LTE performs slightly better at in-plane $2\times$. GSMR3D also achieves the highest lesion sensitivity across all geometries, tied with ResCNN at in-plane $2\times$. At through-plane $3\times$, GSMR3D recovers 97.0\% of the HR-derived lesions and misses 20, compared with 21 for SAINT, 31 for ResCNN, and 38 for the diffusion prior. At $4\times$, where lesion preservation becomes more challenging, GSMR3D retains the highest sensitivity at 91.5\% and misses 57 lesions, compared with 62 for SAINT, 85 for ResCNN, and 89 for the diffusion prior.

GSMR3D also achieves the lowest median enhancing-volume error at isotropic $2\times$ and both through-plane geometries. At through-plane $4\times$, the error is 4.1\%, compared with 5.9\% for SAINT and 6.5\% for the diffusion prior. Together, these results extend the downstream evaluation beyond glioma and show that GSMR3D preserves individual brain metastases and their overall enhancing volume under substantial through-plane degradation. A representative case illustrating lesion preservation across the four degradation geometries is shown in Fig.~\ref{fig:app-qual-bm}.

\subsection{Qualitative results}
\label{app:qual}

\begin{figure*}[p]
\centering
  \includegraphics[width=0.92\linewidth]{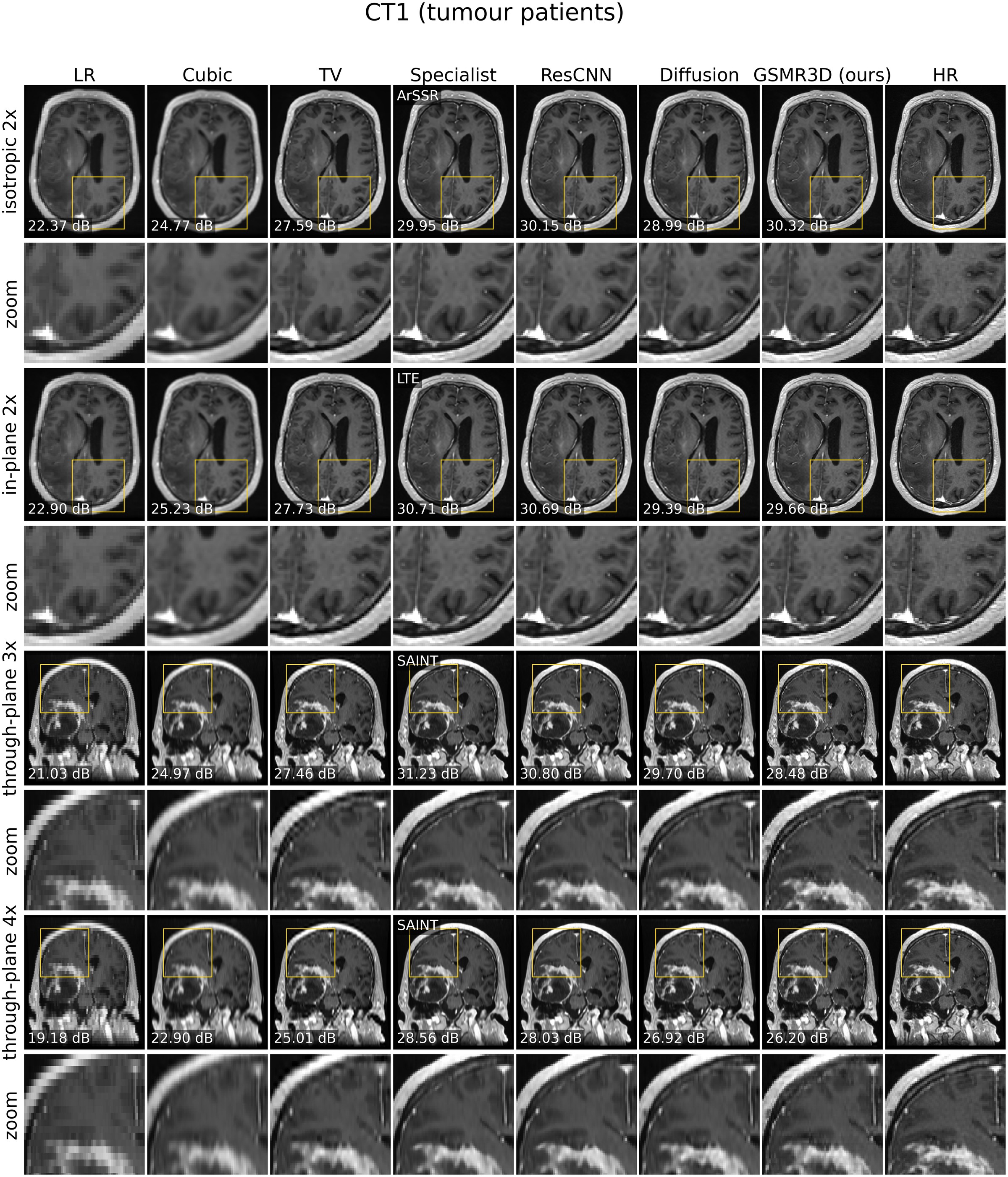}
  \caption{\textbf{Qualitative CT1 comparison across the four degradation geometries} (tumour cohort). Columns show the LR input, cubic interpolation, TV, the geometry-specific specialist (ArSSR, LTE, or SAINT), ResCNN, GSMR3D, and the HR reference. Each reconstruction is accompanied by an enlarged view of the indicated region. Values denote per-case PSNR3D.}
  \label{fig:app-qual-ct1}
\end{figure*}

\begin{figure*}[p]
  \centering
  \includegraphics[width=\linewidth]{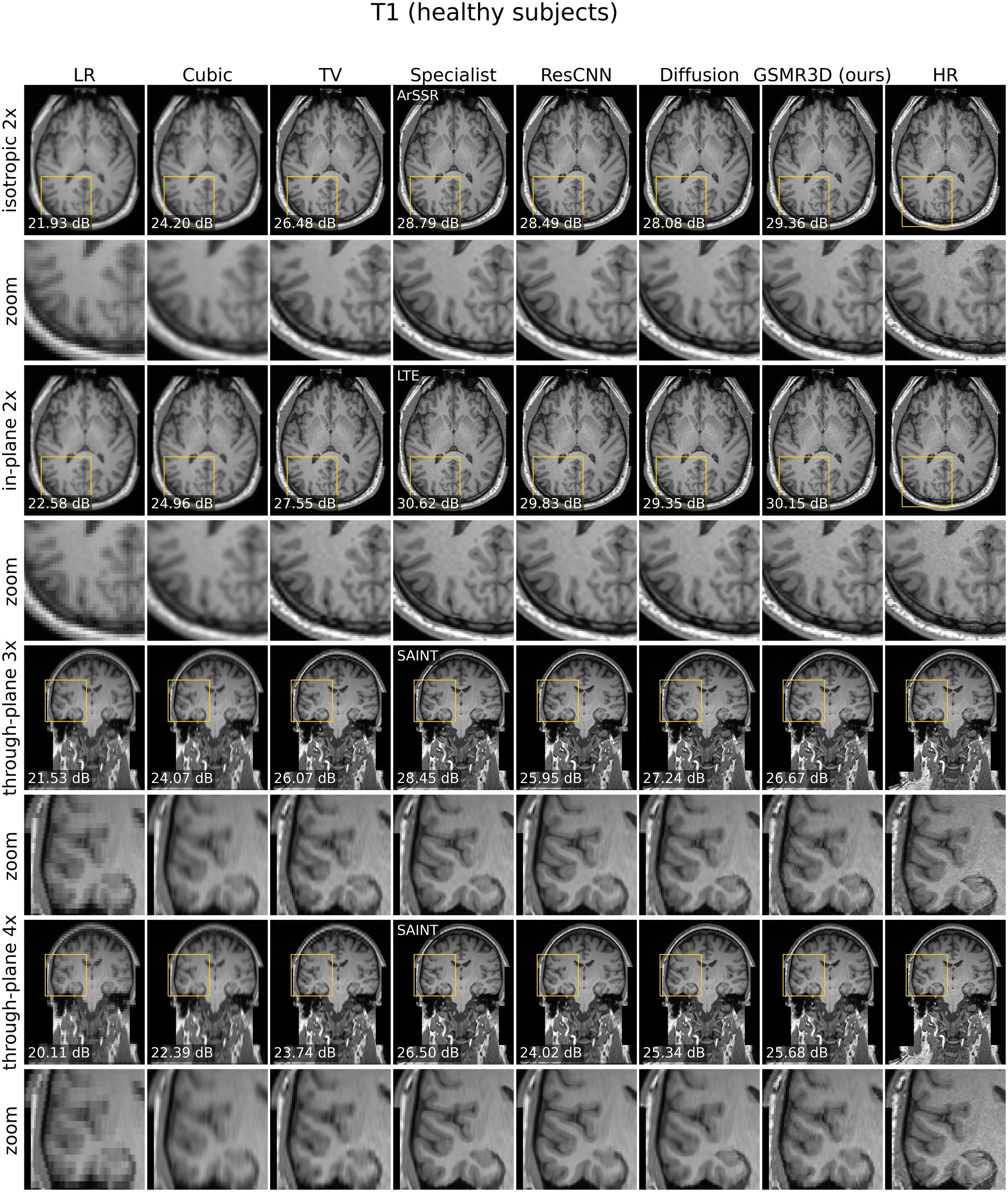}
  \caption{\textbf{Qualitative T1 comparison across the four degradation geometries} (Healthy subject). The layout follows Figure~\ref{fig:app-qual-ct1}.}
  \label{fig:app-qual-t1}
\end{figure*}

Figures~\ref{fig:app-qual-ct1} and \ref{fig:app-qual-t1} show representative CT1 and T1 reconstructions across all four degradation geometries. Cubic interpolation substantially reduces blurring compared with the LR input but does not recover the finer anatomical detail visible in the HR images. At isotropic and in-plane $2\times$, GSMR3D closely reproduces cortical structure and tissue boundaries in both contrasts. The differences between methods become more apparent for through-plane $3\times$ and $4\times$, where the geometry-specific specialist recovers some fine structures more accurately, consistent with its higher PSNR in these settings. Nevertheless, GSMR3D preserves substantially more anatomical detail than interpolation while using the same model across all four geometries.

\begin{figure*}[t]
  \centering
  \includegraphics[width=\linewidth]{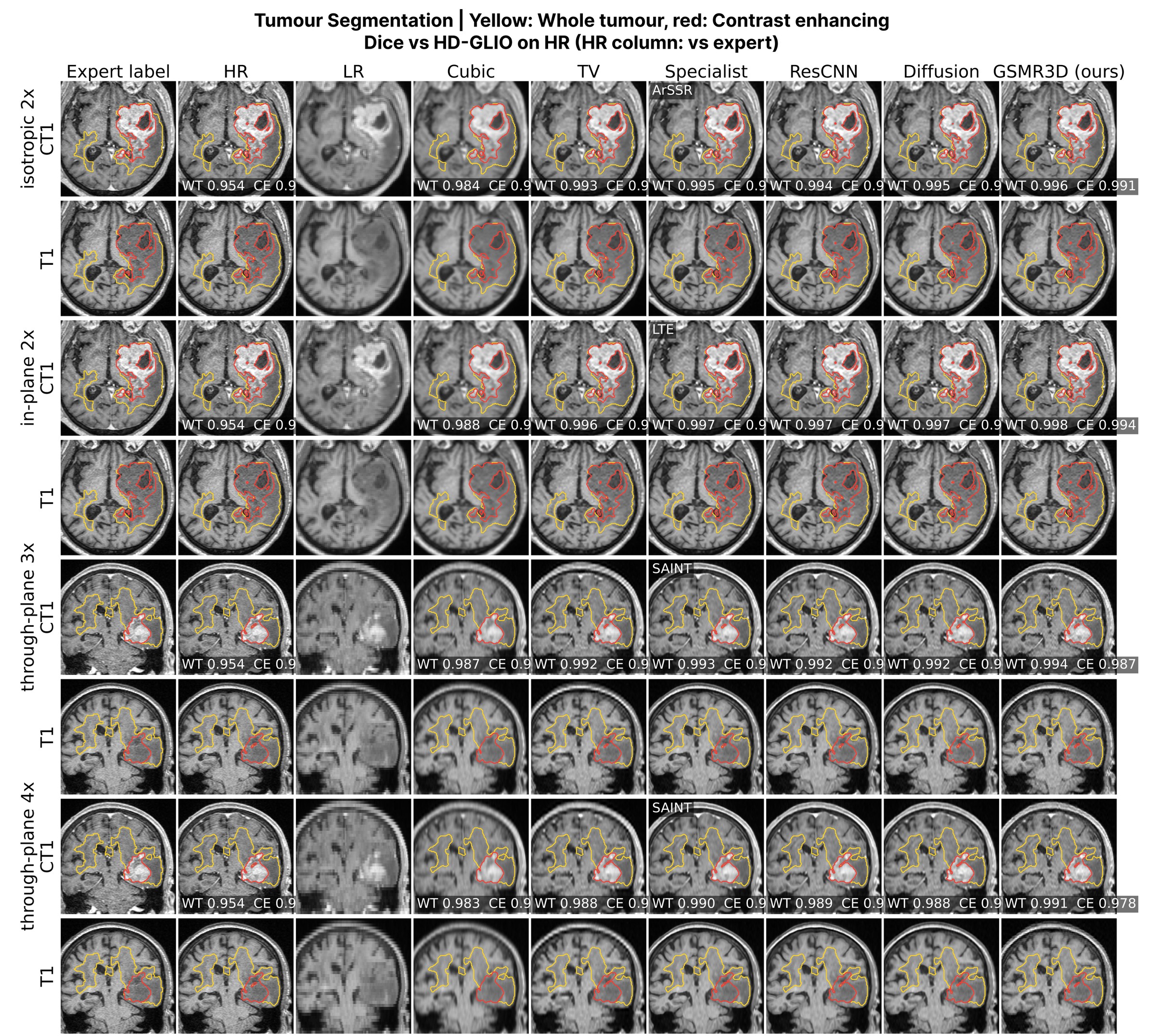}
  \caption{\textbf{HD-GLIO tumour segmentation at different geometries}. Yellow and red contours indicate whole and contrast-enhancing tumour, respectively. Dice scores for reconstructed images measure agreement with the segmentation obtained from the HR volume, while the HR score measures agreement with the expert annotation.}
  \label{fig:app-qual-tumour}
\end{figure*}

Figure~\ref{fig:app-qual-tumour} further illustrates the effect of reconstruction on downstream tumour segmentation under the most severe through-plane $4\times$ degradation. Despite the substantial loss of through-plane information in the LR input, the reconstructed images preserve the overall tumour boundaries and the contrast-enhancing component. In the CT1 example, GSMR3D gives the closest segmentation to that obtained from the HR image, reaching Dice scores of 0.991 for whole tumour and 0.978 for contrast-enhancing tumour. The similarity of the resulting contours to those obtained from HR provides a qualitative counterpart to the segmentation results reported in Table~\ref{tab:seg-full}.

\begin{figure*}[t]
  \centering
  \includegraphics[width=\linewidth]{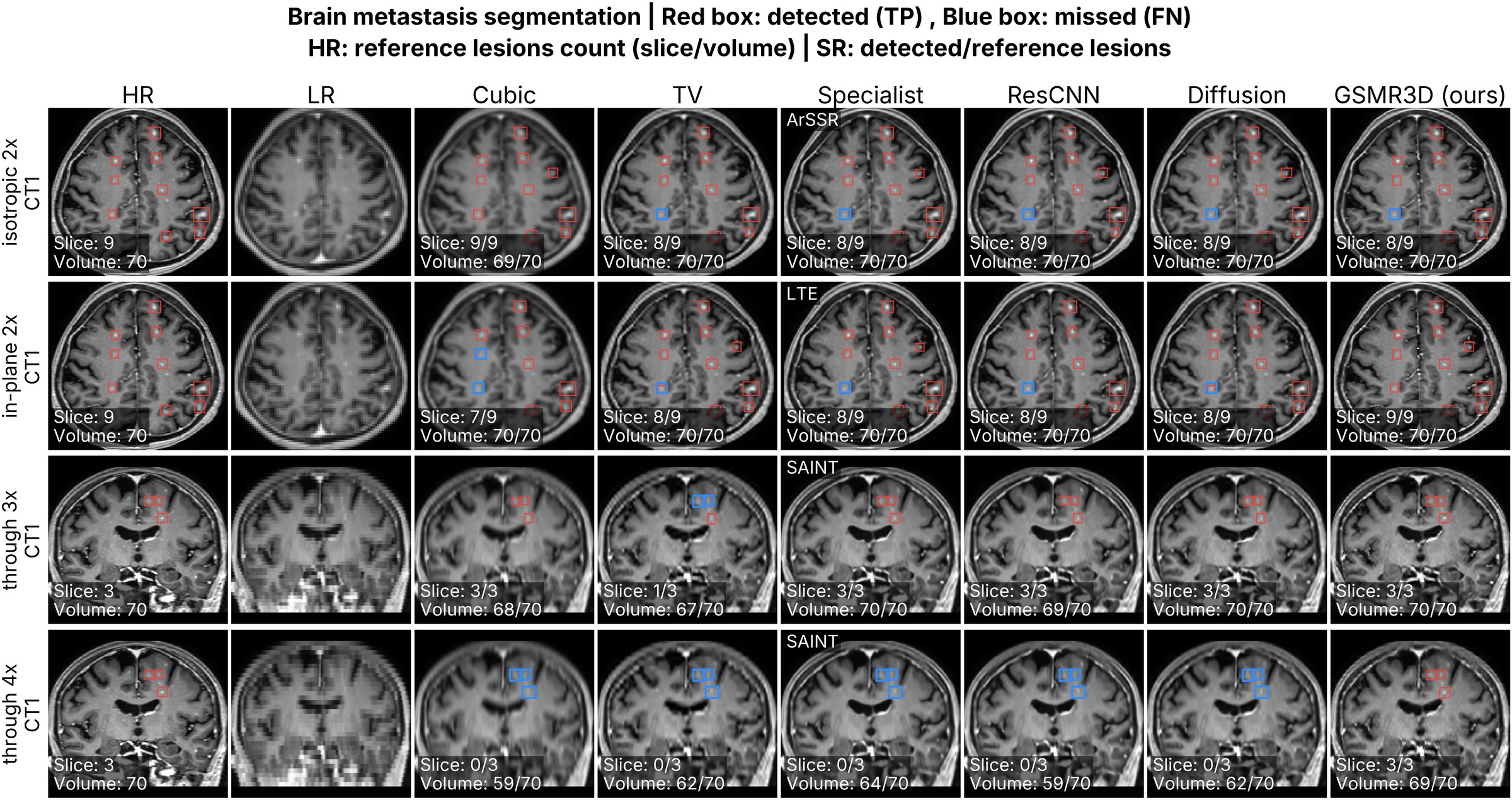}
  \caption{\textbf{Brain-metastasis preservation across acquisition geometries.} Representative CT1 images from a brain-metastasis case with 70 HR-derived lesions. Red boxes indicate detected reference lesions and blue boxes indicate lesions missed on the displayed slice. For the HR image, \emph{Slice} and \emph{Volume} give the number of reference lesions on the displayed slice and in the complete volume. For reconstructed images, these values give the number of detected reference lesions relative to the corresponding HR counts. A lesion missed on the displayed slice may still be detected on a neighbouring slice and therefore counted as detected at the volume level. Volume-level values correspond to the lesion-detection analysis in Table~\ref{tab:bm-seg}.}
  \label{fig:app-qual-bm}
\end{figure*}

Figure~\ref{fig:app-qual-bm} shows a representative brain-metastasis case across the four degradation geometries. The HR volume contains 70 enhancing lesions. GSMR3D detects all 70 lesions at isotropic $2\times$, in-plane $2\times$, and through-plane $3\times$, and 69 of 70 lesions at through-plane $4\times$. Differences between methods are most apparent at $4\times$, where SAINT detects 64 lesions, the diffusion prior and TV detect 62, and ResCNN and cubic interpolation detect 59, compared with 69 for GSMR3D. The slice-level annotations indicate whether each reference lesion visible on the displayed HR slice is also detected on that slice of the SR volume. Because lesion detection is evaluated in 3D, a lesion not detected on the displayed slice may still be detected on an adjacent slice and therefore counted as detected at the volume level.

\subsection{Ablation studies}
\label{sec:ablations}

\paragraph{Protocol.}
All ablations are performed on the CT1 task using a reduced training budget. Dictionary-size experiments are trained for 100 epochs with a global batch size of 16 and $32{,}768$ queries per volume (budget~A). Gather-radius experiments use 100 epochs and a global batch size of 8 (budget~B) because increasing the radius from $R=1$ to $R=2$ increases the neighbourhood from 27 to 125 voxels and substantially increases memory requirements. Each experiment includes a control using the production setting ($K=512$, $R=1$), and results are compared only within the same training budget. We evaluate each checkpoint on 10 held-out validation volumes at eight fixed scale configurations: isotropic $2\times$, $3\times$, and $4\times$, and anisotropic $(1,1,2)$, $(1,1,4)$, $(2,1,1)$, $(2,2,1)$, and $(1,2,4)$. All results use full-volume reconstruction using  the CUDA fused inference kernel.

\paragraph{Dictionary size.}
We first examine how the number of Gaussian prototypes affects SR quality and computational cost. In addition to $K\in\{64,128,256,512,768,1024\}$, we include $K=1$, where a single learned anisotropic Gaussian is shared across all voxels. As shown in Table~\ref{tab:kladder}, increasing the dictionary size generally improves reconstruction quality, with PSNR increasing from 26.72\,dB for a single Gaussian to 28.41\,dB at $K=512$ and 28.84\,dB at $K=1024$. The gains become smaller relative to the additional computational cost beyond $K=512$, while training time and memory increase approximately linearly with dictionary size. We therefore use $K=512$ as a balance between reconstruction quality and computational cost.

\begin{table}[t]
\centering
\caption{\textbf{Dictionary-size ablation} (budget~A). Reconstruction metrics are averaged over ten validation volumes and eight scale configurations. Training time is measured for an $L_1$ training step, and peak memory is reported per GPU.}
\label{tab:kladder}
\adjustbox{max width=\columnwidth}{%
\begin{tabular}{lcccc}
\toprule
Configuration & PSNR$\uparrow$ & SSIM$\uparrow$ & ms/item & Peak mem. \\
\midrule
$K=1$ (single Gaussian) & 26.72 & 0.886 & 24 & 3.8\,GB \\
$K=64$                  & 27.24 & 0.904 & 33 & 5.3\,GB \\
$K=128$                 & 27.92 & 0.910 & 43 & 7.1\,GB \\
$K=256$                 & 28.19 & 0.905 & 65 & 11.6\,GB \\
$K=512$                 & 28.41 & 0.911 & 110 & 20.5\,GB \\
$K=768$                 & 28.49 & 0.913 & 156 & 29.5\,GB \\
$K=1024$                & 28.84 & 0.914 & 202 & 38.4\,GB \\
\bottomrule
\end{tabular}
}
\end{table}

\begin{table}[!tb]
\centering
\scriptsize
\setlength{\tabcolsep}{2pt}
\caption{\textbf{Gather-radius ablation.} Left: the production checkpoint trained with $R=1$ and evaluated using different gather radii. Right: models trained separately with each radius under budget~B. Training uses the autograd (non-fused) gather, so these
costs are independent of the fused inference kernel of Table~\ref{tab:cost}. Evaluation follows Table~\ref{tab:kladder}.}
\label{tab:radius}
\adjustbox{max width=\columnwidth}{%
\begin{tabular}{lcc@{\qquad}lcccc}
\toprule
\multicolumn{3}{c}{Eval-only (production checkpoint)} &
\multicolumn{4}{c}{Trained per radius (budget~B)} \\
\cmidrule(r){1-3}\cmidrule(l){4-7}
Radius & PSNR$\uparrow$ & $\Delta$ &
Radius & PSNR$\uparrow$ & ms/item & Peak mem. \\
\midrule
$R=0$ (1 neighbour)   & 27.01 & $-1.30$ & $R=1$ (27 neighbours)  & \textbf{27.89} & 112 & 10.4\,GB \\
$R=1$ (27 neighbours) & 28.41 & ---     & $R=2$ (125 neighbours) & 27.65 & 422 & 43.0\,GB \\
$R=2$ (125 neighbours)& 28.63 & $+0.22$ & & & & \\
\bottomrule
\end{tabular}}
\end{table}

\begin{table}[!tb]
\centering
\scriptsize
\setlength{\tabcolsep}{2pt}
\caption{\textbf{Inference cost of the fused Gaussian gather.} Query evaluation time (s) and peak memory (GB) for a $64^3\!\rightarrow128^3$ reconstruction containing approximately 2.1M HR queries, evaluated on a single NVIDIA A100 GPU.}
\label{tab:cost}
\adjustbox{max width=\columnwidth}{%
\begin{tabular}{lcccc}
\toprule
 & $K=256$ & $K=512$ & $K=768$ & $K=1024$ \\
\midrule
$R=0$ (1 neighbour)    & 0.40 / 1.0 & 0.45 / 2.0 & 0.49 / 2.2 & 0.54 / 2.5 \\
$R=1$ (27 neighbours)  & 0.67 / 1.0 & 0.98 / 2.0 & 1.29 / 2.2 & 1.60 / 2.5 \\
$R=2$ (125 neighbours) & 1.76 / 1.0 & 3.15 / 2.0 & 4.54 / 2.2 & 5.93 / 2.5 \\
\bottomrule
\end{tabular}}
\end{table}

\paragraph{Gather radius.}
We next examine the size of the LR neighbourhood used by the Gaussian gather Table~\ref{tab:radius}. Since the radius does not introduce additional learned parameters, we first vary it at inference using the production checkpoint trained with $R=1$. Reducing the gather to a single neighbour ($R=0$) decreases PSNR by 1.30\,dB, showing that aggregation across neighbouring LR voxels is important. Expanding the neighbourhood from 27 ($R=1$) to 125 ($R=2$) neighbours gives only a modest gain of 0.22\,dB when applied at inference. We additionally train models directly with each radius under budget~B. Training with $R=2$ does not improve reconstruction quality over $R=1$, while increasing training time by approximately $3.8\times$ and peak memory by $4.1\times$. These results support the $3^3$ neighbourhood used in GSMR3D as a good trade-off between reconstruction quality and computational cost.

\paragraph{Computational cost.}
Table~\ref{tab:cost} measures how dictionary size and gather radius affect inference. Runtime and peak memory increase with both the number of prototypes and the number of neighbouring voxels considered by each query. At the production setting ($K=512$, $R=1$), the fused CUDA implementation requires approximately 0.67\,s per million HR queries on a single NVIDIA A100 GPU, corresponding to approximately 6\,s for a $240^3$ output volume. The fused implementation is $4.8\times$ faster than the equivalent PyTorch implementation at $R=1$ and $7.4\times$ faster at $R=2$. Encoder computation accounts for only approximately 30\,ms, with most inference time therefore spent evaluating the HR queries. Across the dictionary-size sweep, the number of trainable parameters changes only modestly from 4.5M to 5.9M, indicating that the main computational cost of increasing $K$ arises from query evaluation rather than model size.

\end{document}